\documentclass[12pt,a4paper]{article}

\usepackage[margin=1in]{geometry}
\usepackage{setspace}
\usepackage{times}
\usepackage{graphicx}
\usepackage{booktabs}
\usepackage{hyperref}
\usepackage{natbib}
\usepackage{float}
\usepackage{array}

\title{\textbf{AI and Authenticity in Islamic Research: A Critical Evaluation of Generative AI Reliability, Hallucination, and Source Fidelity in Quranic, Hadith, and Fiqh Knowledge}}
\author{Muhammad Sajjad Akbar\textsuperscript{1,*}, Zawar Hussain\textsuperscript{2}, Imran Afzal Khan\textsuperscript{3},\\
Mohammad Polash\textsuperscript{1}, Imdad Ullah\textsuperscript{1}\\[0.6em]
\small \textsuperscript{1}School of Computer Science, The University of Sydney, Australia\\
\small \textsuperscript{2}Macquarie University, Australia\\
\small \textsuperscript{3}The University of New South Wales, Australia\\
\small *Corresponding author: \texttt{sajjad.akr1@gmail.com}}

\date{}

\begin{document}

\maketitle

\begin{abstract}

Generative Artificial Intelligence (AI) is increasingly being used by Muslims to obtain religious guidance, explanations of the Qur'an and Hadith, jurisprudential rulings, and Islamic educational support. However, despite its widespread adoption, limited empirical evidence exists regarding whether current AI systems provide authentic, verifiable, and trustworthy Islamic knowledge suitable for high-trust religious environments. This study addresses this gap by evaluating six leading generative AI systems using a dataset of fifty realistic open-ended Islamic questions spanning Qur'anic interpretation, Hadith explanation, Fiqh reasoning, ethical guidance, pastoral advice, and Madhhab-sensitive topics. Responses were collected under real-world usage conditions from participants in both Australia and the United Kingdom and analysed using a mixed-method evaluation framework incorporating domain accuracy, citation verification, hallucination analysis, jurisprudential consistency, uncertainty handling, source provenance, and comparative geographical analysis.

The evaluation addresses four research questions. First, the study assesses the accuracy and authenticity of AI-generated responses across major Islamic knowledge domains. Second, it investigates the prevalence of hallucinations, fabricated or incomplete citations, unverifiable religious references, and source attribution issues. Third, it compares how different AI systems handle jurisprudential disagreement, Madhhab diversity, uncertainty recognition, and responsible abstention. Finally, it evaluates whether current generative AI systems are sufficiently reliable for high-trust Islamic environments, including religious guidance, Islamic education, and scholarly research.

The findings demonstrate that AI systems achieve their highest reliability in domains characterised by broad scholarly consensus, particularly Qur'anic interpretation and ethical guidance, while substantially lower performance is observed in Fiqh reasoning and Madhhab-sensitive topics. Across all evaluated systems, recurring citation deficiencies were identified, including incomplete Hadith references, missing scholarly attribution, unverifiable religious claims, and occasional hallucinations. Significant differences were also observed in jurisprudential reasoning, citation practices, and uncertainty handling, indicating that model behaviour varies considerably when responding to complex Islamic questions. Furthermore, although responses collected from Australia and the United Kingdom were generally semantically consistent, noticeable variations were observed in supporting references, citation completeness, retrieved sources, and explanatory detail, demonstrating that AI-generated Islamic responses are not entirely deterministic and may be influenced by retrieval environment or geographic access conditions.

Overall, the results indicate that current generative AI systems are valuable as assistive technologies for introductory Islamic learning and information discovery but should not be regarded as authoritative sources for religious rulings or Islamic research without verification against authenticated primary Islamic sources and qualified scholarly expertise. The study provides one of the first comprehensive empirical evaluations of generative AI reliability, citation fidelity, jurisprudential consistency, and geographical response variability within Islamic knowledge, offering practical guidance for researchers, educators, AI developers, and the wider Muslim community.

\end{abstract}


\section{Introduction}

The first point to establish is the scale of generative AI adoption. Generative AI is no longer considered a niche technology. Microsoft’s AI Economy Institute reported that by the second half of 2025, generative AI tools had reached approximately 16.3\% of the world’s population, meaning roughly one in six people were using AI systems for learning, work, or problem solving \cite{microsoft2025globalai}. Similarly, Eurostat reported that 32.7\% of individuals aged 16--74 within the European Union used generative AI tools during 2025 \cite{eurostat2025genai}. In higher education, adoption levels appear even more pronounced. The 2025 Higher Education Policy Institute (HEPI) student survey found that 92\% of students had used some form of AI tool, while 88\% reported using generative AI specifically for assessment-related  \cite{hepi2025survey}. Furthermore, Microsoft’s 2025 education report cited IDC findings indicating that 86\% of educational organizations had already adopted generative AI technologies. Collectively, these statistics demonstrate that AI systems are now embedded within mainstream information-seeking behavior, including contexts involving knowledge-sensitive and educational activities \cite{microsoft2025education}.

However, widespread adoption should not be interpreted as evidence of perfect reliability. On Google DeepMind’s FACTS Grounding benchmark for long-form grounded factuality, leading frontier models still demonstrated notable factual limitations \cite{jacovi2025facts}. Gemini 2.0 Flash Experimental achieved 83.6\%, Gemini 1.5 Flash 82.9\%, Gemini 1.5 Pro 80.0\%, Claude 3.5 Sonnet 79.4\%, GPT-4o 78.8\%, Claude 3.5 Haiku 74.2\%, GPT-4o mini 71.0\%, o1-mini 62.0\%, and o1-preview 61.7\%. OpenAI’s o3/o4-mini system card revealed similar patterns within the SimpleQA benchmark \cite{wei2024measuring}. The o3 model achieved 49\% accuracy, o1 reached 47\%, while o4-mini achieved only 20\%. Corresponding hallucination rates were reported as 51\%, 44\%, and 79\% respectively. These findings do not imply that generative AI systems are unusable; rather, they indicate that even advanced models continue to produce materially incorrect responses frequently enough to create concerns within high-stakes domains.

The issue becomes increasingly concerning when considering citation fidelity and expert-domain applications. A 2025 \textit{JMIR Mental Health} study examining GPT-4o-generated literature review citations found that 19.9\% of generated citations were entirely fabricated \cite{linardon2025influence}. Among the citations that were genuine, 45.4\% contained substantive errors. The study concluded that nearly two-thirds of citations were either fabricated or inaccurate. Similar reliability concerns have also emerged in legal-domain AI research. Stanford’s legal-AI studies reported that earlier investigations found general-purpose chatbots hallucinated on legal queries between 58\% and 82\% of the time \cite{magesh2025hallucination}. Their later benchmarking study further demonstrated that legal research tools continued to hallucinate in at least one out of every six evaluation queries. These findings are highly relevant for Islamic knowledge research because Islamic scholarship depends heavily upon accurate attribution, source traceability, contextual interpretation, and jurisprudential precision rather than merely producing fluent text.

The third point, and perhaps the most critical for this study’s problem statement, is that Islam-specific benchmarking now empirically confirms substantial performance gaps among large language models. The 2026 IslamicMMLU benchmark evaluated 26 LLMs using 10,013 questions spanning Quranic studies, Hadith, and Fiqh. Overall accuracy scores ranged from 39.8\% to 93.8\%. Selected results included Gemini 3 Flash at 93.8\%, GPT-5 at 89.9\%, Claude Sonnet 4.5 at 86.2\%, Claude 3.7 Sonnet at 82.3\%, GPT-4o at 77.3\%, GPT-4.1 at 75.6\%, GPT-4 at 59.6\%, and GPT-3.5-turbo at 39.8\%. Importantly, these evaluations were conducted using multiple-choice questions, which are generally easier than open-ended fatwa-style reasoning or interpretive dialogue. The benchmark authors explicitly noted that large language models are increasingly being consulted for Islamic knowledge despite the previous absence of comprehensive Islamic-domain benchmarks \cite{abdelaal2026islamicmmlu}.

Additional Islamic-domain studies further reinforce the same conclusion. Within the FiqhQA benchmark, designed for school-specific Islamic rulings, researchers observed significant performance variation across models, languages, and madhhabs. GPT-4o demonstrated comparatively stronger accuracy, whereas Gemini and Fanar showed stronger abstention behavior \cite{atif2025sacred}. This distinction is important because refusing uncertain responses may be safer than confidently generating incorrect religious rulings. Similarly, the IslamTrust benchmark, which evaluated alignment with consensus-based Islamic values, reported that the highest-performing model achieved only 66.5\% overall alignment \cite{lahmar2025islamtrust}. Research on Islamic inheritance reasoning demonstrated additional disparities: o3 achieved 93.4\%, Gemini 2.5 scored 90.6\%, GPT-4.5 achieved 74\%, while ALLaM, Fanar, LLaMA, and Mistral all scored below 50\% \cite{bouchekif2025assessing}. These findings collectively demonstrate that generative AI performance within Islamic contexts remains imperfect and highly sensitive to task complexity, language, reasoning depth, and school-of-thought distinctions.

A careful methodological consideration is also necessary when comparing AI systems. It is generally more rigorous to evaluate foundational model families rather than consumer-facing applications because many public AI tools operate as wrappers, routed systems, or interfaces layered over the same underlying model architectures. Consequently, the most defensible research narrative is not necessarily identifying which application is ``best overall,'' but rather examining how leading foundational model families perform on factuality, reasoning, and Islamic-knowledge benchmarks under controlled evaluation settings. The benchmarks discussed above provide a strong empirical foundation for such analysis.

\section{Motivation}

The rapid emergence of generative Artificial Intelligence has fundamentally changed how people access information, including religious knowledge. Millions of users now rely on large language models to obtain explanations, guidance, and answers that were traditionally sought from books, scholars, educational institutions, or trusted online repositories. Within the Muslim community, this shift has significant implications because Islamic knowledge is founded upon authenticated primary sources, recognised scholarly methodologies, and established principles of interpretation. Consequently, understanding how accurately and responsibly generative AI represents Islamic knowledge has become an important research problem rather than merely a technological curiosity.

Throughout Islamic history, Muslims have embraced beneficial scientific and technological developments while simultaneously emphasising the importance of verification, authenticity, and evidence before accepting information. The Qur'an itself instructs believers to verify information before acting upon it, highlighting the importance of critical evaluation rather than unquestioning acceptance. As generative AI becomes increasingly integrated into education, research, and everyday information seeking, it is therefore essential that its capabilities and limitations be examined using rigorous empirical methods rather than assumptions, anecdotal experiences, or isolated examples.

This need is particularly important because Islamic information differs from many other knowledge domains. The correctness of an answer depends not only on factual accuracy but also on authentic citation of the Qur'an and Hadith, recognition of scholarly consensus and legitimate differences of opinion, transparency regarding uncertainty, and appropriate representation of jurisprudential reasoning. An answer that appears fluent and persuasive may nevertheless contain incomplete references, oversimplified interpretations, or unsupported religious claims. Such limitations can have greater consequences in religious contexts where individuals may rely on AI-generated responses when making personal, educational, or religious decisions.

The motivation of this study is therefore not to determine whether generative AI should or should not be used for Islamic learning, but to provide the Muslim community, educators, researchers, and developers with objective, evidence-based knowledge regarding its current capabilities and limitations. By systematically evaluating multiple leading AI systems across diverse Islamic knowledge domains, this research enables users to make informed decisions about when AI-generated responses may be useful, when additional scholarly verification is required, and where greater caution should be exercised. Such evidence contributes to the responsible adoption of emerging technologies while preserving the authenticity, transparency, and scholarly integrity that are fundamental to Islamic knowledge.

\begin{figure*}[htbp]
    \centering
    \includegraphics[width=\textwidth]{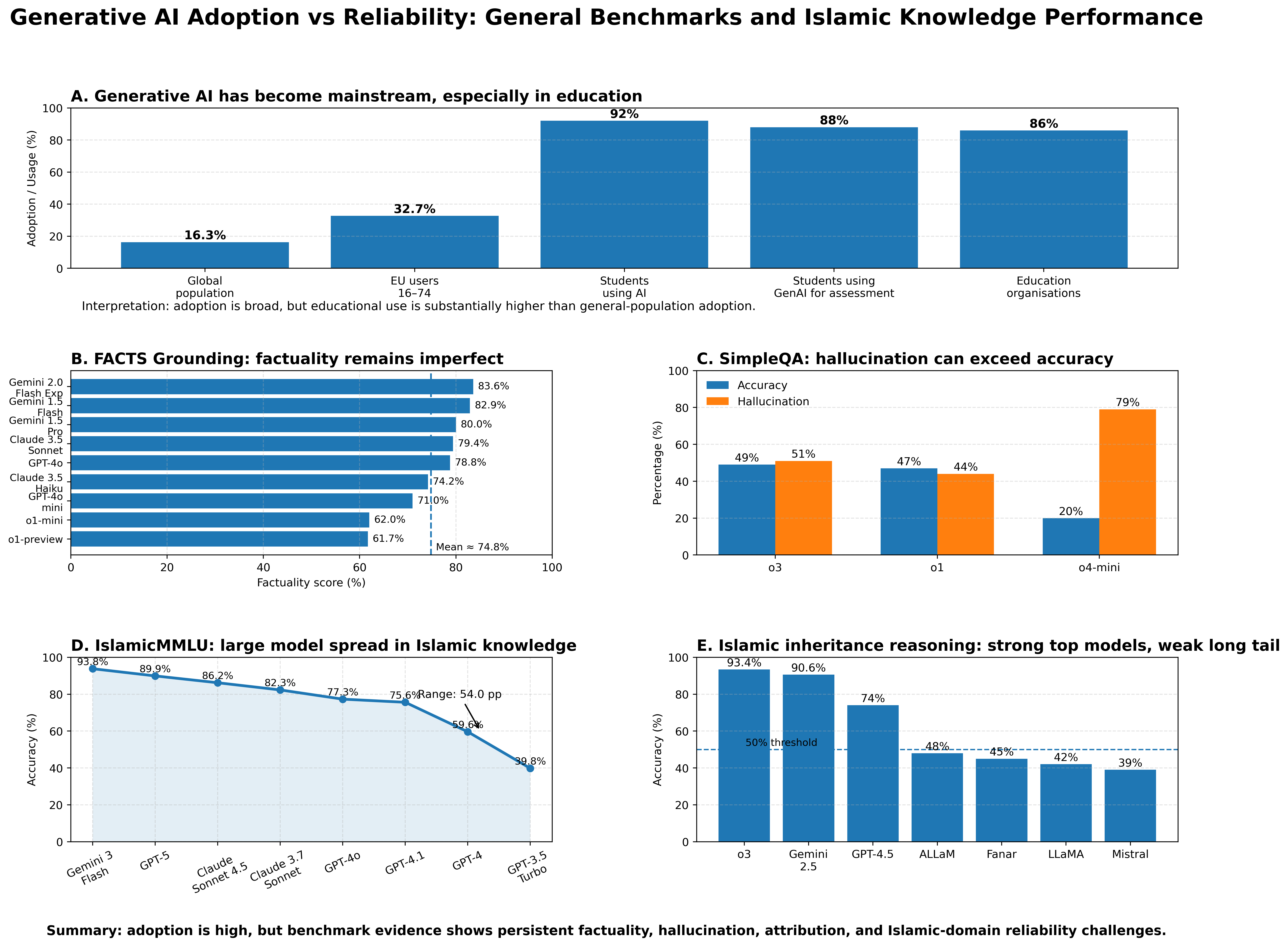}
    \caption{Generative AI adoption and reliability challenges across general factuality benchmarks and Islamic-domain evaluation tasks.}
    \label{fig:ai_islamic_reliability}
\end{figure*}

Figure~\ref{fig:ai_islamic_reliability} summarises the current landscape of generative AI adoption and reliability using evidence from recent benchmark studies. Panel A demonstrates that generative AI has achieved widespread adoption, particularly within education. While approximately 16.3\% of the global population and 32.7\% of EU users (aged 16--74) report using generative AI, adoption is substantially higher in educational settings, where 92\% of students report using AI, 88\% use it specifically for assessments, and 86\% of educational organisations have adopted AI technologies. These statistics indicate that AI has rapidly become an integral component of modern teaching and learning.

Despite this widespread adoption, Panels B and C show that current AI systems remain imperfect in terms of factual reliability. FACTS Grounding benchmark results in Panel B report factuality scores ranging from approximately 61.7\% to 83.6\%, with an average of only 74.8\%, demonstrating that even state-of-the-art models do not consistently generate factually correct information. Furthermore, the SimpleQA benchmark in Panel C highlights the continuing challenge of hallucinations. For example, the o3 model achieved 49\% factual accuracy while exhibiting a 51\% hallucination rate, and GPT-4o Mini produced only 20\% accuracy alongside a 79\% hallucination rate, illustrating that confidently generated but incorrect information remains a significant limitation.

Panels D and E extend this analysis to Islamic knowledge benchmarks. On the IslamicMMLU benchmark, model accuracy varies considerably from 93.8\% for Gemini 2.0 Flash to only 39.8\% for GPT-3.5 Turbo, representing a performance spread of approximately 54 percentage points. A similar trend is observed for Islamic inheritance reasoning, where o3 and Gemini 2.5 achieve 93.4\% and 90.6\% accuracy, respectively, whereas several open-source and smaller models fall below the 50\% performance threshold. These results demonstrate that although leading AI systems show promising performance on certain Islamic tasks, reliability differs substantially across models and specialised domains. Collectively, the benchmark evidence indicates that widespread adoption of generative AI should not be interpreted as evidence of consistent factual correctness or religious reliability, thereby motivating the need for the comprehensive evaluation conducted in this study.

\section{Problem Statement}
The widespread uptake of generative AI has transformed how people seek information, including in educational, professional, and belief-oriented contexts. By late 2025, generative AI tools were already being used by roughly one in six people globally, by nearly one-third of people in the EU, and by the overwhelming majority of surveyed higher-education students. This scale of adoption means AI systems are no longer experimental aids; they are active intermediaries in how users learn, verify, and act on information. 
However, existing research shows that strong fluency does not guarantee factual reliability. On established factuality benchmarks, leading models still produce non-trivial rates of error and hallucination, while studies in research and legal settings have shown fabricated citations, inaccurate references, and confidently stated falsehoods. These limitations are especially problematic in domains where meaning depends not only on surface correctness, but on authentic sourcing, interpretive precision, and responsible abstention when uncertainty is high. 
This concern is particularly acute in the Islamic domain. Islamic knowledge is grounded in authenticated sources such as the Qur’an and Hadith, shaped by established methods of interpretation, and, in many cases, mediated by legitimate differences across schools of thought. Recent Islamic-domain benchmarks already show substantial variability in LLM performance across Quran, Hadith, Fiqh, inheritance reasoning, and Islamic-values alignment, with even strong frontier models falling meaningfully short of perfect accuracy or alignment. Yet despite growing public use of AI for religious questions, there remains insufficient empirical evidence about how reliably these systems provide authentic Islamic answers in real user-facing settings, especially when questions require source citation, handling of juristic disagreement, or recognition of uncertainty. 
Accordingly, the core problem is not simply whether AI can answer Islamic questions, but whether it can do so authentically, verifiably, and responsibly. This creates an urgent research need to evaluate AI-generated Islamic responses for factual correctness, source authenticity, citation validity, school-of-thought sensitivity, and consistency across tools. Without such evaluation, there is a real risk that fluent but unauthenticated outputs will be mistaken for trustworthy religious guidance, potentially amplifying misunderstanding at scale. Following are the research questions: 

\begin{enumerate}
    \item[\textbf{RQ1.}] 
    How accurately and authentically do leading generative AI systems answer Islamic questions across the domains of Qur’an, Hadith, Fiqh, and Islamic reasoning?

    \item[\textbf{RQ2.}] 
    To what extent do AI-generated Islamic responses contain hallucinations, fabricated citations, inaccurate religious references, or unverifiable sources?

    \item[\textbf{RQ3.}] 
    How consistently do different generative AI models handle Islamic jurisprudential issues, including school-of-thought differences (\textit{madhhabs}), uncertainty recognition, and responsible abstention?

    \item[\textbf{RQ4.}] 
    How suitable are current generative AI systems for high-trust Islamic information environments such as religious guidance, education, and Islamic research in terms of factual correctness, source authenticity, and interpretive reliability?
\end{enumerate}

\section{Data Collection}

The survey instrument was intentionally designed to evaluate all four research questions. Questions covering Qur'anic interpretation, Hadith, Fiqh, ethics, and pastoral guidance supported the evaluation of response accuracy (RQ1). Jurisprudential and school-of-thought-sensitive questions enabled assessment of consistency, uncertainty handling, and madhhab awareness (RQ3). Citation quality, source attribution, and reference verification were analysed to identify hallucinations and authenticity issues (RQ2). Collectively, these evaluation dimensions informed the assessment of the suitability of generative AI for high-trust Islamic environments (RQ4).
This study adopted a survey-based empirical data collection methodology to evaluate the authenticity, reliability, factual correctness, and interpretive consistency of AI-generated Islamic responses. The research design was intentionally developed to simulate realistic public interaction with generative AI systems rather than artificial benchmark-only evaluation environments. The primary objective was to observe how ordinary users interact with modern AI systems when seeking Islamic knowledge, guidance, clarification, or religious interpretation.

To achieve this objective, a structured survey instrument consisting of \textbf{50 open-ended Islamic questions} was developed. The survey was divided into \textbf{10 independent sets}, with each set containing \textbf{5 questions}. The questions were intentionally designed to reflect authentic user-facing Islamic information-seeking behavior commonly observed in online communities, educational settings, social media discussions, and AI-assisted search interactions. In the analysis, for the quality purpose, we will establish conclusion based on 5 sets, where we received complete responses.

The survey questions covered three major Islamic knowledge domains:

\begin{itemize}
    \item \textbf{Qur’anic Knowledge and Interpretation}
    \item \textbf{Hadith and Prophetic Teachings}
    \item \textbf{Fiqh and Practical Islamic Jurisprudence}
\end{itemize}

In addition to purely factual religious questions, the survey also incorporated:
\begin{itemize}
    \item emotionally contextualized questions,
    \item ethical dilemmas,
    \item school-of-thought-sensitive issues,
    \item repentance and spiritual guidance scenarios,
    \item and practical real-life religious decision-making situations.
\end{itemize}

This design was intentional because modern users increasingly interact with generative AI systems not only for factual retrieval, but also for interpretive guidance, emotional reassurance, ethical clarification, and jurisprudential understanding.

The 10 survey sets are summarized below.

\subsection{Survey Question Sets}

\subsubsection*{Set 1: Honesty, Intentions, and Prayer}
\begin{enumerate}
    \item What does the Qur’an say about honesty and truthfulness?
    \item Provide a Hadith about intentions and explain it.
    \item What should a person do if they miss a prayer?
    \item I feel guilty for past sins, what does Islam say?
    \item Why is honesty emphasized in Islam?
\end{enumerate}

\subsubsection*{Set 2: Taqwa, Kindness, and Scholarly Differences}
\begin{enumerate}
    \item Explain the concept of Taqwa in the Qur’an.
    \item What did the Prophet say about kindness?
    \item Is music allowed in Islam? Explain different opinions.
    \item I am confused between scholars, what should I do?
    \item What is the role of intention in actions?
\end{enumerate}

\subsubsection*{Set 3: Patience, Charity, and Fasting}
\begin{enumerate}
    \item What does the Qur’an say about patience?
    \item Provide a Hadith about charity.
    \item What breaks wudu?
    \item I accidentally ate during fasting, what now?
    \item Why is charity important?
\end{enumerate}

\subsubsection*{Set 4: Justice, Knowledge, and Ramadan}
\begin{enumerate}
    \item What does the Qur’an say about justice?
    \item What is the Hadith on seeking knowledge?
    \item What are the rules of fasting in Ramadan?
    \item I cannot wake up for Fajr, what should I do?
    \item Why is knowledge important in Islam?
\end{enumerate}

\subsubsection*{Set 5: Prophet Musa, Anger, and Prayer}
\begin{enumerate}
    \item Explain the story of Prophet Musa and its lessons.
    \item What did the Prophet say about anger?
    \item What invalidates prayer?
    \item I struggle with anger, what does Islam advise?
    \item What is the role of patience?
\end{enumerate}

\subsubsection*{Set 6: Zakat, Forgiveness, and Charity}
\begin{enumerate}
    \item What does the Qur’an say about charity (Zakat)?
    \item Provide a Hadith about forgiveness.
    \item What are the conditions of Zakat?
    \item I want to give charity, how much should I give?
    \item Why is charity emphasized in Islam?
\end{enumerate}

\subsubsection*{Set 7: Judgment, Neighbours, and Accountability}
\begin{enumerate}
    \item How does the Qur’an describe the Day of Judgment?
    \item What did the Prophet say about neighbours?
    \item What is the ruling on tattoos?
    \item I face discrimination, what does Islam say?
    \item What is the concept of accountability?
\end{enumerate}

\subsubsection*{Set 8: Surah Al-Fatiha, Sincerity, and Riba}
\begin{enumerate}
    \item What is the meaning of Surah Al-Fatiha?
    \item Provide a Hadith about sincerity.
    \item What is the ruling on interest (riba)?
    \item I earn interest income, what should I do?
    \item Why is interest prohibited in Islam?
\end{enumerate}

\subsubsection*{Set 9: Forgiveness, Ethics, and Hajj}
\begin{enumerate}
    \item What does the Qur’an say about forgiveness?
    \item What is the Hadith about good character?
    \item What are the requirements for Hajj?
    \item I want to repent sincerely, how?
    \item What is the role of ethics in Islam?
\end{enumerate}

\subsubsection*{Set 10: Tawakkul, Lying, and Halal/Haram}
\begin{enumerate}
    \item Explain the concept of Tawakkul.
    \item What did the Prophet say about lying?
    \item What is the ruling on insurance?
    \item I am unsure if my job is halal, how should I assess it?
    \item What is the philosophy of halal and haram?
\end{enumerate}

\subsection{Participant Interaction with AI Systems}

Participants were intentionally given unrestricted freedom regarding the generative AI systems they could use. No limitations were imposed on:
\begin{itemize}
    \item AI platform selection,
    \item subscription level,
    \item model family,
    \item or interaction style.
\end{itemize}

Participants could therefore use any publicly available generative AI system, including ChatGPT, Gemini, Claude, Copilot, DeepSeek, and other AI assistants. This unrestricted design was methodologically important because the study aimed to evaluate real-world AI-assisted religious information-seeking behavior rather than controlled laboratory-only benchmark performance.

Allowing participants to freely choose AI tools also enabled comparative analysis across different model ecosystems, retrieval systems, and generative architectures. Since participants naturally selected different AI platforms, the collected responses reflected realistic variation in:
\begin{itemize}
    \item factual accuracy,
    \item citation behavior,
    \item hallucination tendencies,
    \item interpretive reasoning,
    \item and jurisprudential handling.
\end{itemize}

\subsection{Geographical Distribution}

The survey was distributed in two geographical regions:
\begin{itemize}
    \item Australia
    \item United Kingdom
\end{itemize}

The inclusion of geographically distinct participant groups was intended to explore whether AI-generated Islamic responses differ across regions. This consideration was important because modern AI systems may utilize:
\begin{itemize}
    \item geographically distributed infrastructure,
    \item regional model routing,
    \item localization mechanisms,
    \item location-aware retrieval pipelines,
    \item or region-specific moderation policies.
\end{itemize}

Collecting responses from both Australia and the United Kingdom therefore enabled exploratory investigation into whether geographical context influences the authenticity, consistency, or interpretive behavior of AI-generated Islamic answers.

\subsection{Data Analysis Dimensions}

Each collected response was preserved and systematically analyzed using both qualitative and quantitative evaluation dimensions. The analysis focused on:

\begin{itemize}
    \item factual correctness,
    \item authenticity of Qur’anic references,
    \item authenticity of Hadith citations,
    \item source traceability,
    \item fabricated or hallucinated content,
    \item jurisprudential consistency,
    \item handling of madhhab differences,
    \item uncertainty recognition,
    \item abstention behavior,
    \item interpretive reliability,
    \item and consistency across AI systems.
\end{itemize}

The resulting dataset therefore represents a large-scale, realistic corpus of AI-assisted Islamic information-seeking interactions generated under unconstrained user conditions. Unlike purely synthetic benchmark environments, this dataset captures how generative AI systems are practically used by real individuals when seeking Islamic understanding, religious clarification, and guidance-oriented responses.

Table~\ref{tab:comparative_similarity} summarises the overall similarity observed across the five survey sets analysed in this study. The table provides an overview of the thematic focus of each set together with the relative similarity level and the primary observations that guided the subsequent detailed analysis.

\begin{table}[h]
\centering
\caption{Comparative Similarity Analysis of the Survey Sets}
\begin{tabular}{|c|p{4cm}|p{2.5cm}|p{5.5cm}|}
\hline
\textbf{Sets} & \textbf{Theme} & \textbf{Similarity Level} & \textbf{Comment} \\
\hline
1 & Honesty, intentions, prayer, guilt & High & Same core verses/hadith and similar moral framing \\
\hline
2 & Taqwa, kindness, music, scholar confusion & Medium & Same themes but different depth/style \\
\hline
3 & Patience, charity, wudu, fasting & Medium-low & Australia much longer and more jurisprudential \\
\hline
4 & Justice, knowledge, fasting, Fajr & Low-medium & UK tab appears incomplete/short \\
\hline
5 & Musa, anger, invalid prayer, patience & Medium & Same Islamic themes, different tool style \\
\hline

\end{tabular}
\label{tab:comparative_similarity}
\end{table}

Table~\ref{tab:ai_fingerprint_patterns} summarises the characteristic response patterns identified across the evaluated AI systems. These behavioural fingerprints assisted in distinguishing stylistic differences independently of factual correctness.

\begin{table}[H]
\centering
\caption{Observed AI Tool Fingerprint Patterns}
\begin{tabular}{|p{3cm}|p{10cm}|}
\hline
\textbf{AI Tool} & \textbf{Fingerprint Pattern} \\
\hline
ChatGPT & Balanced tone, ``simple summary,'' ``bottom line,'' ``if you want'' follow-ups, and moderate reference usage \\
\hline
Claude & Longer academic style, structured headings, philosophical explanation, and stronger organisational structure \\
\hline
Gemini & Educational and explanatory style, often balanced but generally less detailed than Claude \\
\hline
Copilot & Shorter, cleaner, and more direct answer style \\
\hline
DeepSeek & More categorical and traditional ruling-focused responses with stronger certainty \\
\hline
Dola AI & Structured but sometimes generic responses; often provides reference-like outputs \\
\hline
\end{tabular}
\label{tab:ai_fingerprint_patterns}
\end{table}

Table~\ref{tab:authenticity} summarises the major authenticity and citation issues identified during the evaluation. These observations formed the basis for the hallucination and citation reliability analysis presented later in the Findings section.

\begin{table}[H]
\centering
\caption{Observed Authenticity and Citation Issues in AI-Generated Islamic Responses}
\begin{tabular}{|p{5cm}|p{8cm}|}
\hline
\textbf{Issue Type} & \textbf{Observation} \\
\hline
Missing exact hadith numbers & Common \\
\hline
General references only & Common in ChatGPT and Copilot responses \\
\hline
Strong claims without source verification & Present across multiple responses \\
\hline
``Reported in Sahih Muslim/Bukhari'' without exact citation & Frequent \\
\hline
Mixed scholarly views without named scholars & Common \\
\hline
Possible overconfidence & Observed particularly in DeepSeek-style insurance responses \\

\hline
\end{tabular}
\label{tab:authenticity}
\end{table}

\section{Methodology}
This study adopts a mixed-method comparative evaluation framework integrating thematic analysis, source validation, comparative survey analysis, and benchmark-inspired Islamic AI evaluation methodologies. The methodological design was conceptually informed by Islamic AI benchmarking approaches such as IslamicMMLU, where Islamic knowledge is categorised into multiple reasoning domains including Qur’anic interpretation, Hadith analysis, Fiqh reasoning, jurisprudential interpretation, and ethical or pastoral guidance. However, unlike traditional benchmark systems that primarily rely on structured multiple-choice evaluation, the present study focuses on open-ended AI-generated Islamic responses collected through real-world user interaction.

Hallucination and references issued found, many answers mention Qur’an or Hadith references but do not provide full source verification, Arabic text, hadith number, grading, or school-specific context.

\subsection{Ethical and Foundational Islamic Guidance (Set 1)}

\begin{figure}[H]
\centering
\includegraphics[width=\textwidth]{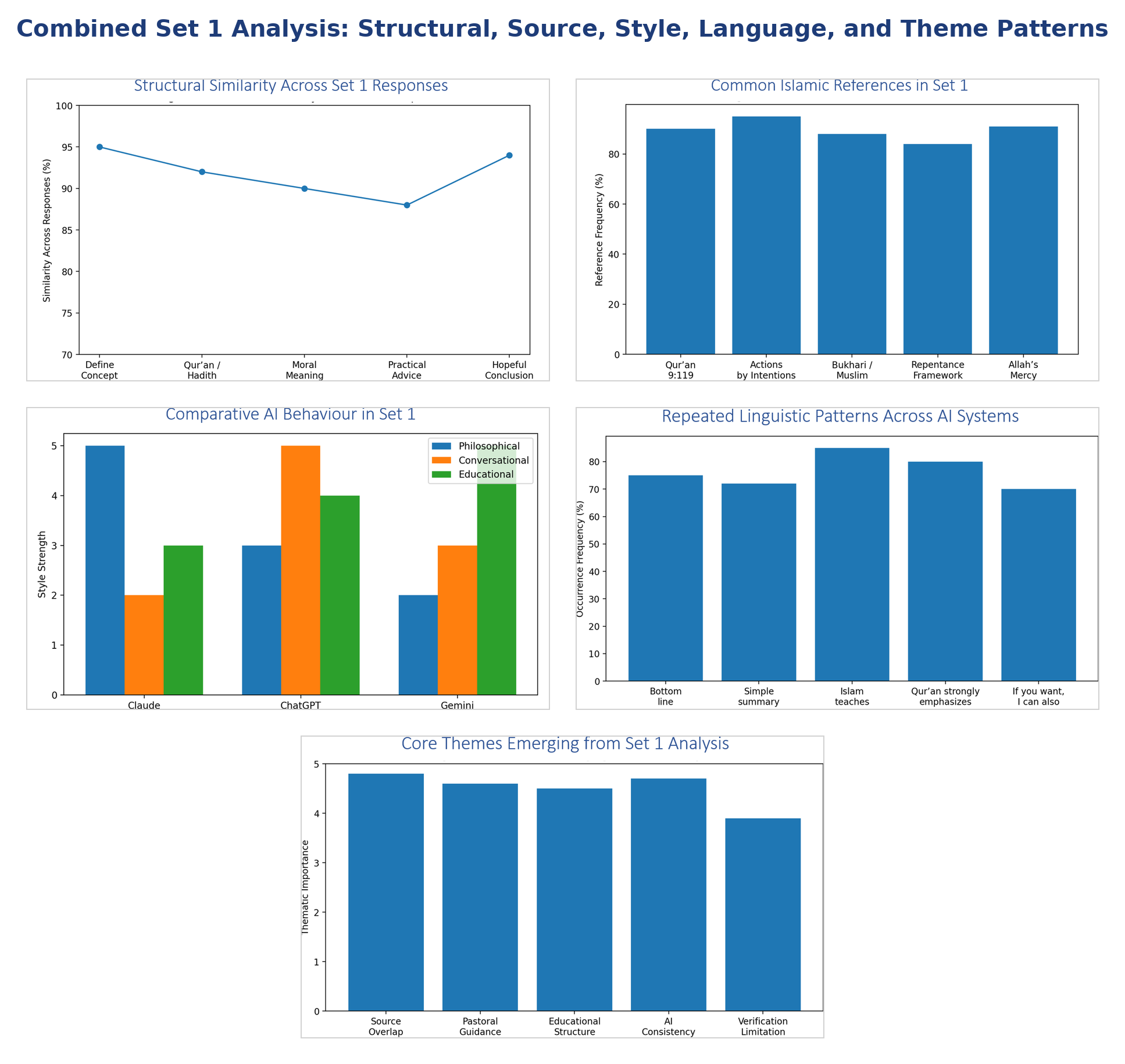}
\caption{Combined analysis of Set 1 responses, showing structural similarity, common Islamic references, AI behavioural patterns, repeated linguistic patterns, and core thematic findings.}
\label{fig:combined_set1_analysis}
\end{figure}

The combined Figure~\ref{fig:combined_set1_analysis} integrates five interconnected analytical dimensions identified in Set 1 concerning honesty, intentions, missed prayer, guilt, and repentance. Rather than presenting isolated observations, the combined image demonstrates how structural organisation, Islamic source usage, AI stylistic behaviour, repeated linguistic framing, and thematic consistency collectively contributed to the very high similarity observed across the UK and Australian datasets.

The first component of the combined figure, ``Structural Similarity Across Set 1 Responses,'' demonstrates that most AI-generated answers followed a highly standardised educational structure. Across different tools and geographical datasets, responses repeatedly followed a similar logical sequence involving definition of the Islamic concept, provision of Qur’anic or Hadith evidence, explanation of moral meaning, practical advice, and finally a hopeful conclusion. This structural consistency indicates that modern generative AI systems have learned a highly stable Islamic educational-response pattern from overlapping religious educational corpora. The graph therefore highlights the strong instructional regularity present across AI-generated Islamic answers.

The second component, ``Common Islamic References in Set 1,'' illustrates the repeated use of similar Qur’anic verses, Hadith narrations, and repentance concepts across the datasets. Frequently repeated references included Qur’an 9:119, the Hadith ``Actions are judged by intentions,'' references to Sahih al-Bukhari and Sahih Muslim, and repentance-related concepts involving mercy and forgiveness. The significance of this graph lies in its demonstration of substantial source overlap across AI systems. The repeated appearance of the same references strongly suggests reliance upon highly indexed Islamic educational websites and standardised online religious discourse.

The third component, ``Comparative AI Behaviour in Set 1,'' compares the stylistic tendencies of Claude, ChatGPT, and Gemini. The graph demonstrates that Claude responses were generally more philosophical, reflective, and essay-oriented. By contrast, ChatGPT responses were more conversational, balanced, and practically educational, while Gemini responses demonstrated stronger educational-summary behaviour with simplified explanation and concise instructional framing. This figure is significant because it shows that although the systems shared similar Islamic source material, they still exhibited distinguishable stylistic fingerprints.

The fourth component, ``Repeated Linguistic Patterns Across AI Systems,'' highlights commonly repeated rhetorical phrases identified throughout the AI-generated responses. Expressions such as ``Bottom line,'' ``Simple summary,'' ``Islam teaches,'' and ``If you want, I can also \ldots'' appeared repeatedly across different tools and datasets. The graph demonstrates strong linguistic overlap and suggests that transformer-based AI systems have learned highly similar summarisation and instructional behaviours from overlapping Islamic educational content.

The final component, ``Core Themes Emerging from Set 1 Analysis,'' integrates the broader thematic conclusions identified throughout the comparative analysis. The graph demonstrates that source overlap, pastoral guidance, educational structuring, AI consistency, and verification limitations collectively contributed to the high similarity observed across the UK and Australian datasets. Importantly, the figure shows that the observed similarity was not solely due to direct source reuse, but also due to shared instructional framing, overlapping educational ecosystems, and standardised Islamic online discourse structures.

Overall, the combined figure demonstrates that Set 1 produced the strongest cross-dataset consistency because the questions primarily involved mainstream Islamic ethics, repentance, sincerity, honesty, and moral guidance, which are comparatively less jurisprudentially disputed than later fiqh-oriented tabs. Consequently, AI systems converged toward highly similar educational and pastoral response patterns across multiple tools and geographic datasets.

\subsection{Jurisprudential Disagreement and Scholarly Diversity (Set 2)}

\begin{figure}[H]
\centering
\includegraphics[width=\textwidth]{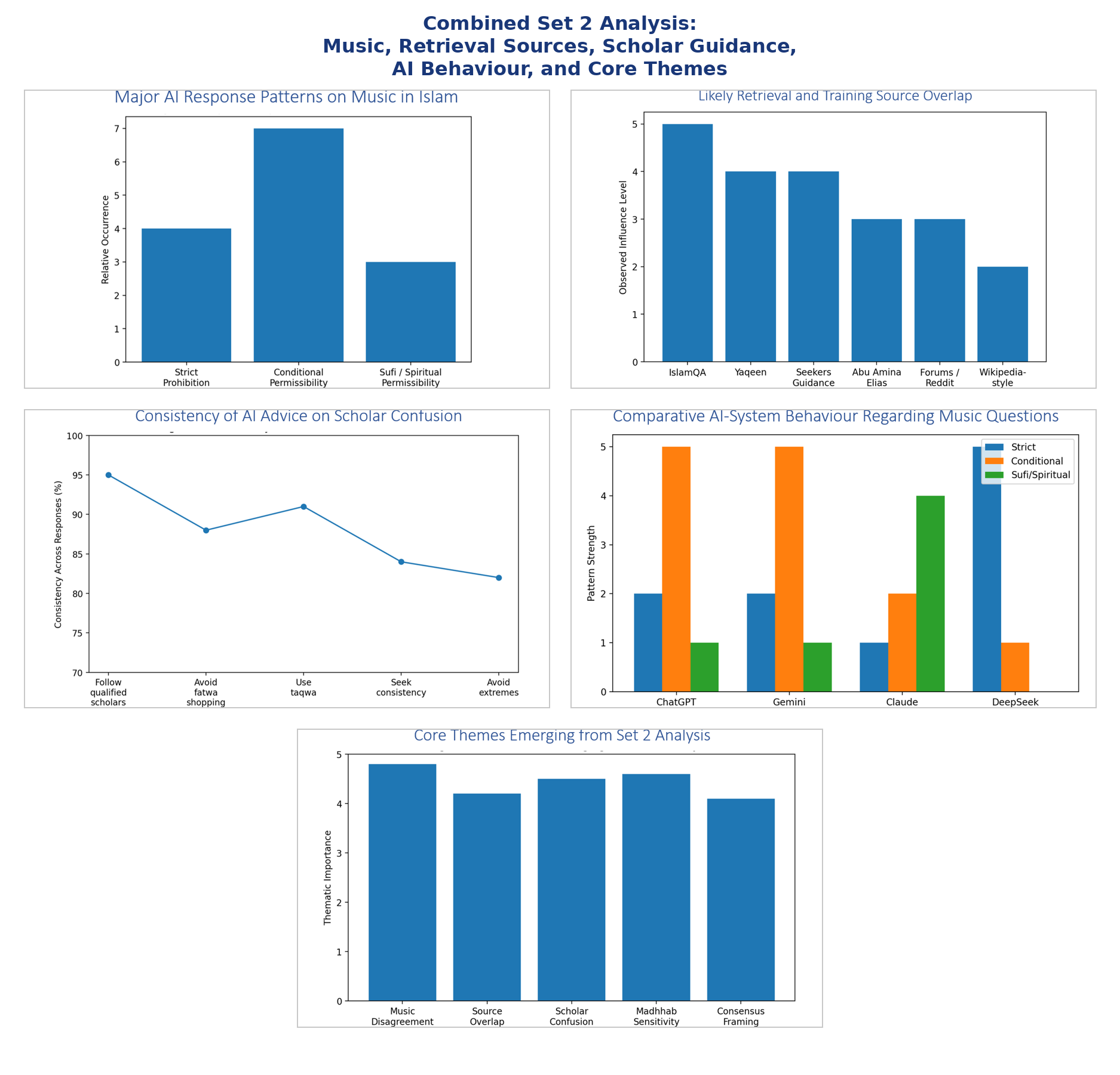}
\caption{Combined analysis of Set 2 responses showing music-related jurisprudential patterns, retrieval-source overlap, scholar-confusion guidance consistency, comparative AI-system behaviour, and core thematic findings across the UK and Australian datasets.}
\label{fig:combined_set2_analysis}
\end{figure}

The combined Figure~\ref{fig:combined_set2_analysis} for Set 2 integrates five interconnected analytical dimensions concerning taqwa, kindness, music, and scholar confusion. Rather than presenting isolated visual findings, the combined image demonstrates how jurisprudential disagreement, retrieval-source overlap, AI stylistic behaviour, moderation-oriented guidance, and thematic consistency collectively shaped the medium similarity level observed across the UK and Australian datasets.

The first component of the combined figure, ``Major AI Response Patterns on Music in Islam,'' demonstrates that AI systems did not converge toward a single Islamic ruling regarding music. Instead, three dominant response patterns emerged across the datasets: strict prohibition, conditional permissibility, and Sufi or spiritual permissibility. The figure illustrates that conditional permissibility appeared most frequently, while stricter prohibition-oriented responses were more strongly associated with systems such as DeepSeek. The graph highlights the direct impact of Islamic jurisprudential disagreement (ikhtilaf) on AI-generated religious guidance and demonstrates that AI responses are heavily influenced by the interpretive traditions embedded within their retrieval and training ecosystems.

The second component, Likely Retrieval and Training Source Overlap,'' illustrates the probable influence of major online Islamic educational platforms on AI-generated responses. The repeated appearance of concepts such as lahw al-hadith,'' ma‘azif,'' difference of opinion,'' and ``context matters'' strongly suggests retrieval overlap with highly indexed Islamic websites including IslamQA, Yaqeen Institute, SeekersGuidance, Abu Amina Elias, Reddit-style Islamic discussions, and comparative educational summaries. The graph demonstrates that modern AI systems likely depend upon overlapping Islamic educational ecosystems that significantly shape doctrinal framing, interpretive boundaries, and linguistic consistency within generated responses.

The third component, ``Consistency of AI Advice on Scholar Confusion,'' demonstrates the high consistency observed when AI systems responded to questions involving conflicting scholarly opinions. Across multiple tools and datasets, AI systems repeatedly recommended following qualified scholars, avoiding fatwa shopping, maintaining taqwa, seeking consistency, and avoiding extremes. The figure highlights that despite jurisprudential variation in substantive rulings, AI systems strongly converged toward moderation-oriented Islamic educational narratives. This consistency suggests that contemporary online Islamic discourse has become highly standardised in its pastoral and advisory framing.

The fourth component, ``Comparative AI-System Behaviour Regarding Music Questions,'' compares how different AI systems approached music-related Islamic questions. The figure demonstrates that ChatGPT and Gemini more frequently adopted conditional permissibility frameworks, balancing scholarly disagreement with contextual interpretation. Claude responses showed stronger reflective and spiritual framing, occasionally incorporating Sufi-oriented perspectives and philosophical explanation. By contrast, DeepSeek more frequently adopted stricter prohibition-oriented reasoning with stronger certainty and reduced interpretive nuance. This graph therefore highlights that although AI systems share overlapping source ecosystems, they still exhibit identifiable jurisprudential and stylistic fingerprints.

The final component, ``Core Themes Emerging from Set 2 Analysis,'' integrates the broader thematic conclusions identified throughout the comparative analysis. The graph demonstrates that music disagreement, source overlap, scholar confusion, madhhab sensitivity, and consensus-oriented framing were strongly interconnected throughout the datasets. Importantly, the figure shows that authenticity challenges in Islamic AI systems extend beyond simple hallucination problems. Instead, the findings indicate deeper concerns involving interpretive governance, jurisprudential transparency, doctrinal framing, and responsible handling of scholarly disagreement.

Overall, the combined figure demonstrates that Set 2 produced lower similarity than Set 1 because questions involving music and conflicting scholarly opinions inherently require jurisprudential interpretation rather than straightforward moral guidance. Consequently, AI systems exhibited greater variation in doctrinal positioning, interpretive framing, and certainty behaviour across different models and geographic datasets.

\subsection{Ritual Worship and Fiqh Complexity (Set 3)}

\begin{figure}[H]
\centering
\includegraphics[width=\textwidth]{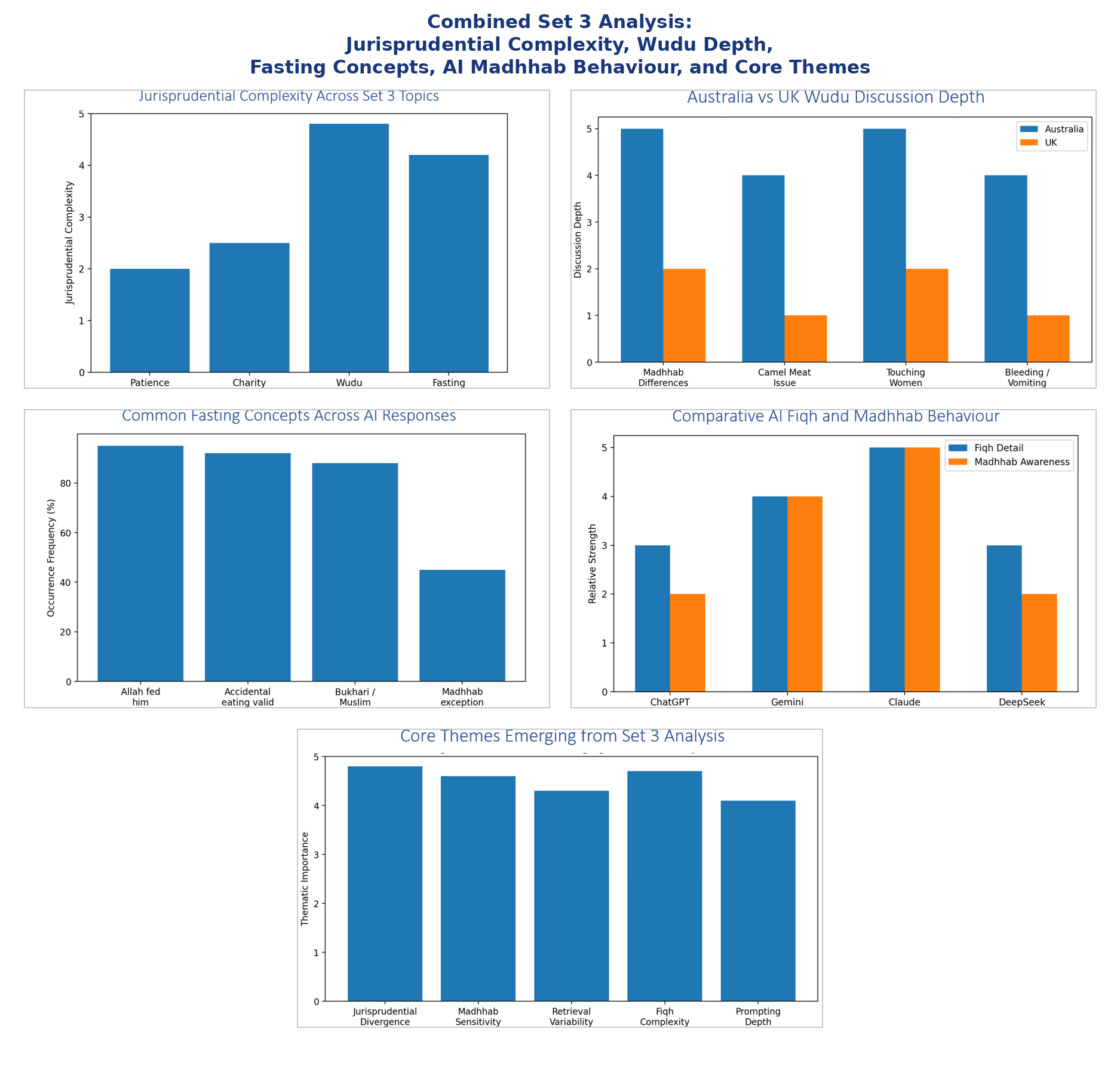}
\caption{Combined analysis of Set 3 responses showing jurisprudential complexity, comparative wudu discussion depth, fasting-related concepts, AI fiqh and madhhab behaviour, and the major thematic findings emerging across the UK and Australian datasets.}
\label{fig:combined_set3_analysis}
\end{figure}

The combined Figure~\ref{fig:combined_set3_analysis} for Set 3 integrates five interconnected analytical dimensions concerning patience, charity, wudu, and fasting. Rather than presenting isolated visual findings, the combined image demonstrates how jurisprudential complexity, madhhab sensitivity, fiqh-oriented depth, retrieval variability, and comparative AI-system behaviour collectively contributed to the lower similarity levels observed across the UK and Australian datasets.

The first component of the combined figure, ``Jurisprudential Complexity Across Set 3 Topics,'' demonstrates that wudu and fasting generated substantially greater fiqh complexity than patience and charity. The graph illustrates that questions involving ritual purity and fasting required more technical jurisprudential interpretation, including madhhab-sensitive rulings and contextual legal reasoning. By contrast, patience and charity produced lower jurisprudential complexity because they primarily involved ethical guidance rather than detailed legal interpretation. This figure therefore highlights how fiqh-oriented questions naturally increase variability within AI-generated Islamic responses.

The second component, ``Australia vs UK Wudu Discussion Depth,'' compares the level of fiqh-oriented discussion identified across the Australian and UK datasets. The figure demonstrates that Australian responses consistently contained more detailed treatment of madhhab differences, camel meat rulings, touching women, and bleeding or vomiting issues. UK responses, by contrast, were generally shorter and focused primarily on broad principles rather than comparative jurisprudential analysis. The graph suggests that either Australian participants prompted AI systems for greater detail or that certain tools such as Claude and Gemini retrieved more fiqh-oriented Islamic content during response generation.

The third component, Common Fasting Concepts Across AI Responses,'' summarises the most frequently repeated fasting-related concepts identified throughout the datasets. The graph demonstrates strong overlap in explanations involving Allah fed him,'' accidental eating rulings, and references to Sahih al-Bukhari and Sahih Muslim. However, madhhab-specific fasting exceptions appeared significantly less frequently across responses. This figure is important because it suggests that retrieval variability, rather than direct hallucination, explains many of the observed differences in fasting-related fiqh discussion.

The fourth component, ``Comparative AI Fiqh and Madhhab Behaviour,'' compares how different AI systems handled fiqh detail and madhhab-aware reasoning. The graph demonstrates that Claude and Gemini generally produced stronger fiqh detail and greater awareness of madhhab differences, while ChatGPT and DeepSeek more frequently generated simplified jurisprudential explanations with reduced comparative nuance. This figure therefore highlights that AI systems exhibit distinguishable jurisprudential and interpretive fingerprints even when responding to highly similar Islamic questions.

The final component, ``Core Themes Emerging from Set 3 Analysis,'' integrates the broader thematic conclusions identified throughout the comparative analysis. The graph demonstrates that jurisprudential divergence, madhhab sensitivity, retrieval variability, fiqh complexity, and prompting depth were strongly interconnected throughout the datasets. Collectively, these factors contributed to the lower similarity levels observed in Set 3 compared with earlier sets focused primarily on moral and ethical guidance.

Overall, the combined figure demonstrates that Set 3 marked a transition from relatively standardised Islamic educational responses toward more technically jurisprudential and madhhab-sensitive discourse. As questions became more fiqh-oriented, AI systems increasingly diverged in terms of detail, interpretive framing, madhhab awareness, and retrieval behaviour, thereby producing greater variability across both models and geographic datasets.

\subsection{Behavioural and Motivational Guidance (Set 4)}

\begin{figure}[H]
\centering
\includegraphics[width=\textwidth]{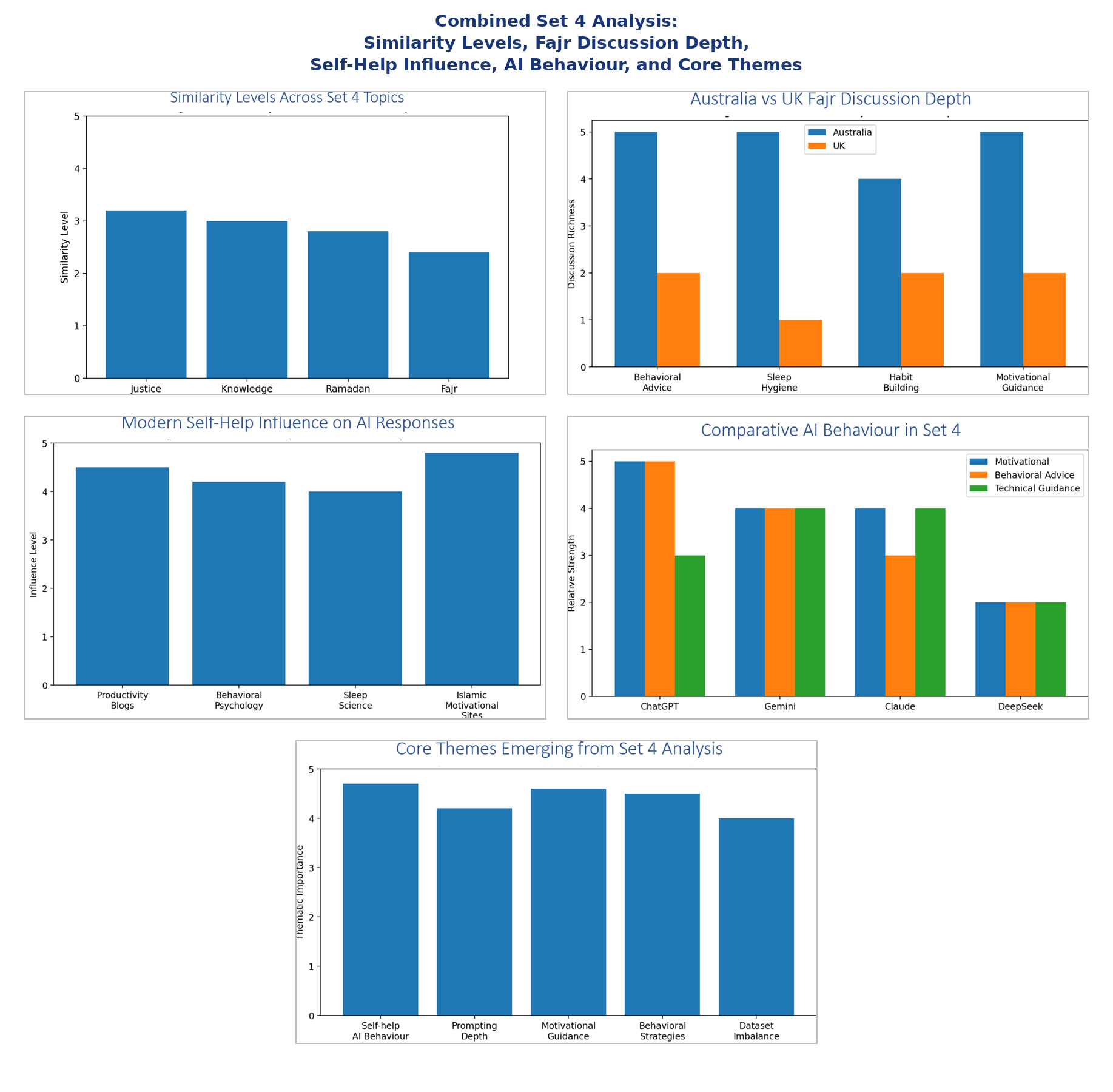}
\caption{Combined analysis of Set 4 responses showing similarity levels, comparative Fajr discussion depth, modern self-help influence, AI-system behavioural patterns, and the major thematic findings emerging across the UK and Australian datasets.}
\label{fig:combined_set4_analysis}
\end{figure}

The combined Figure~\ref{fig:combined_set4_analysis} for Set 4 integrates five interconnected analytical dimensions concerning justice, knowledge, Ramadan, and Fajr prayer. Rather than presenting isolated graphical observations, the combined image demonstrates how dataset imbalance, motivational framing, behavioural guidance, modern self-help influence, and comparative AI-system behaviour collectively contributed to the lower similarity levels observed across the UK and Australian datasets.

The first component of the combined figure, ``Similarity Levels Across Set 4 Topics,'' demonstrates that similarity levels across Set 4 were lower than earlier tabs. The graph illustrates that the Australian dataset consistently contained richer and more detailed responses, while the UK dataset included shorter and partially incomplete answers. This imbalance reduced structural similarity across the datasets and increased variation in educational framing, motivational tone, and practical guidance.

The second component, ``Australia vs UK Fajr Discussion Depth,'' compares the richness of behavioural and motivational discussion surrounding Fajr prayer across both datasets. The figure demonstrates that Australian responses provided substantially more detailed advice regarding sleep hygiene, alarm strategies, habit-building, behavioural consistency, and motivational reinforcement. By contrast, UK responses were generally shorter and less behaviourally developed. This suggests that Australian participants either prompted AI systems for greater practical detail or interacted with tools producing more expansive motivational guidance.

The third component, ``Modern Self-Help Influence on AI Responses,'' highlights the strong influence of contemporary self-help and productivity ecosystems on AI-generated Islamic responses. The graph demonstrates that concepts commonly associated with productivity blogs, behavioural psychology, sleep science, and Islamic motivational websites appeared frequently throughout Fajr-related discussions. This finding suggests that AI systems increasingly merge Islamic educational content with modern behavioural optimisation discourse, thereby producing hybrid religious and self-help response styles.

The fourth component, ``Comparative AI Behaviour in Set 4,'' compares how different AI systems handled motivational framing, behavioural advice, and technical guidance. The figure demonstrates that ChatGPT and Gemini generated stronger motivational and behavioural-oriented responses, often focusing on habit-building and practical self-improvement strategies. Claude produced comparatively more reflective and educational explanations, while DeepSeek responses were shorter and less behaviourally detailed. The graph therefore highlights clear stylistic and instructional differences across AI systems despite addressing similar Islamic questions.

The final component, ``Core Themes Emerging from Set 4 Analysis,'' integrates the broader thematic conclusions identified throughout the comparative analysis. The graph demonstrates that self-help AI behaviour, prompting depth, motivational guidance, behavioural strategies, and dataset imbalance were strongly interconnected across the datasets. Collectively, these factors contributed to the reduced similarity levels observed in Set 4 compared with earlier tabs focused more heavily on standardised ethical or theological content.

Overall, the combined figure demonstrates that Set 4 reflects a transition toward modern hybrid Islamic-self-help discourse within generative AI systems. Rather than relying solely on classical religious explanation, many responses incorporated behavioural optimisation, motivational framing, productivity-style advice, and psychological guidance. Consequently, AI-generated Islamic responses increasingly reflected both religious educational structures and contemporary self-improvement ecosystems, thereby producing greater stylistic variability across tools and geographic datasets.

\subsection{Narrative and Story-Based Islamic Responses (Set 5)}

\begin{figure}[H]
\centering
\includegraphics[width=\textwidth]{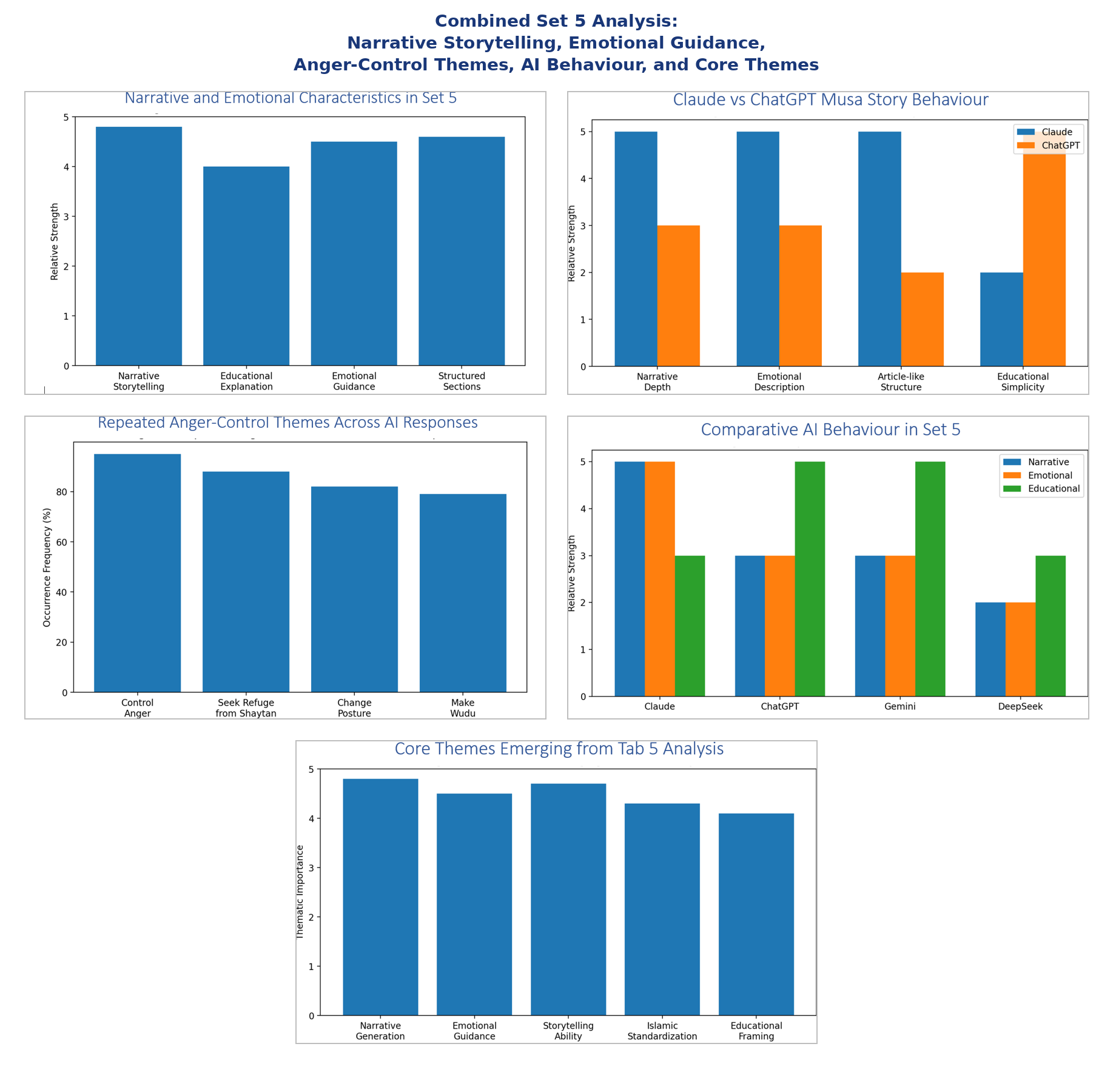}
\caption{Combined analysis of Set 5 responses showing narrative and emotional characteristics, Musa (AS) storytelling behaviour, repeated anger-control themes, comparative AI-system behaviour, and the major thematic findings emerging across the UK and Australian datasets.}
\label{fig:combined_set5_analysis}
\end{figure}

The combined Figure~\ref{fig:combined_set5_analysis} for Set 5 integrates five interconnected analytical dimensions concerning Musa (AS), anger, invalid prayer, and patience. Rather than presenting isolated graphical findings, the combined image demonstrates how narrative storytelling, emotional guidance, educational structuring, standardized Islamic advice, and comparative AI-system behaviour collectively contributed to the medium similarity level observed across the UK and Australian datasets.

The first component of the combined figure, ``Narrative and Emotional Characteristics in Set 5,'' demonstrates that the responses strongly emphasized narrative storytelling, emotional explanation, and structured guidance. Questions involving Musa (AS) particularly encouraged AI systems to generate descriptive and emotionally engaging responses rather than purely factual explanations. The graph therefore highlights the increasing role of narrative-oriented Islamic educational discourse within generative AI systems.

The second component, ``Claude vs ChatGPT Musa Story Behaviour,'' compares how Claude and ChatGPT handled Musa (AS)-related narrative questions. The figure demonstrates that Claude responses were substantially more narrative, emotionally descriptive, reflective, and article-like in structure. By contrast, ChatGPT responses were comparatively simpler, more concise, and more educationally structured. This difference represents a strong stylistic AI fingerprint, illustrating how individual AI systems can generate distinct forms of Islamic storytelling despite relying upon overlapping source material.

The third component, ``Repeated Anger-Control Themes Across AI Responses,'' illustrates the strong overlap in standardized Islamic anger-management advice across the datasets. Frequently repeated concepts included controlling anger, seeking refuge from Shaytan, changing posture, and making wudu. The graph demonstrates that these themes appeared consistently across multiple AI systems, suggesting heavy reliance on widely circulated Islamic educational websites and commonly repeated online religious guidance.

The fourth component, ``Comparative AI Behaviour in Set 5,'' compares how different AI systems handled narrative generation, emotional guidance, and educational explanation. The figure demonstrates that Claude exhibited the strongest narrative and emotional storytelling behaviour, while ChatGPT and Gemini produced more educationally structured and simplified guidance. DeepSeek responses were comparatively shorter and less narratively detailed. This graph therefore highlights clear stylistic and pedagogical differences across AI systems within Islamic educational contexts.

The final component, ``Core Themes Emerging from Set 5 Analysis,'' integrates the broader thematic conclusions identified throughout the comparative analysis. The figure demonstrates that narrative generation, emotional guidance, storytelling ability, Islamic educational standardization, and educational framing were strongly interconnected throughout the datasets. Collectively, these factors contributed to the medium similarity level observed in Set 5, where AI systems balanced standardized Islamic advice with varying narrative and emotional styles.

Overall, the combined figure demonstrates that Set 5 reflects the strong narrative capabilities of modern generative AI systems within Islamic discourse. Unlike highly jurisprudential tabs, the Musa (AS) and anger-related questions encouraged emotionally expressive, story-driven, and motivational responses. Consequently, AI systems converged around standardized Islamic ethical guidance while still exhibiting distinct stylistic fingerprints in narrative structure, emotional framing, and educational presentation.

\subsection{Mapping Evaluation Criteria to Research Questions}
\label{sec:method_rq_mapping}

To ensure alignment between the research design, analysis process, and findings, the evaluation criteria were mapped directly to the four research questions. This mapping was used to organise the coding of AI-generated responses and to structure the presentation of results in Section~6. The purpose of this step was to ensure that each research question was supported by clear analytical criteria and corresponding evidence.

\begin{table}[htbp]
\centering
\caption{Mapping of Research Questions to Evaluation Criteria and Evidence}
\label{tab:rq_method_mapping}
\renewcommand{\arraystretch}{1.25}
\begin{tabular}{|p{3.2cm}|p{5.8cm}|p{5.8cm}|}
\hline
\textbf{Research Question} & \textbf{Evaluation Criteria} & \textbf{Evidence Produced} \\
\hline

\textbf{RQ1} &
Domain accuracy, authenticity of responses, and comparative risk across Qur'an, Hadith, Fiqh, ethics, pastoral guidance, and Madhhab sensitivity. &
Performance radar chart and Islamic knowledge-domain risk assessment. \\
\hline

\textbf{RQ2} &
Citation verification, hallucination analysis, source attribution, fabricated or incomplete references, and unverifiable religious claims. &
Hallucination issue table, incomplete reference examples, citation reliability analysis, and source validation findings. \\
\hline

\textbf{RQ3} &
Fiqh consistency, Madhhab awareness, recognition of scholarly disagreement, uncertainty handling, and responsible abstention. &
Comparative thematic evaluation across AI systems, including fluency, citation completeness, fiqh consistency, and uncertainty handling. \\
\hline

\textbf{RQ4} &
Suitability for high-trust Islamic environments, source authenticity, interpretive reliability, knowledge provenance, and real-world reliability risks. &
Knowledge provenance analysis, source ecosystem mapping, Tajweed error example, and device-based source variation example. \\
\hline

\end{tabular}
\end{table}

As shown in Table~\ref{tab:rq_method_mapping}, each research question was operationalised through specific evaluation criteria. \textbf{RQ1} was assessed by comparing domain-level accuracy and risk across different areas of Islamic knowledge. \textbf{RQ2} was evaluated through citation verification and hallucination analysis, with particular attention to missing Hadith references, unverifiable claims, and incomplete source attribution. \textbf{RQ3} focused on jurisprudential reasoning, including whether AI systems recognised Madhhab differences, handled uncertainty responsibly, and avoided overconfident rulings in disputed matters. Finally, \textbf{RQ4} synthesised the preceding analyses to assess whether current generative AI systems are suitable for high-trust Islamic information environments such as religious guidance, education, and Islamic research.

This mapping provides the methodological bridge between the collected AI-generated responses and the findings reported in Section~6. It ensures that the analysis moves beyond general comparison of AI systems and instead evaluates their performance against clearly defined religious, evidential, and interpretive criteria.

\section{Findings}

This section presents the findings obtained from the evaluation of leading generative AI systems across a comprehensive set of Islamic knowledge tasks. To provide a structured analysis, the findings are organised according to the study's four research questions. First, Section~\ref{sec:rq1_domains} evaluates the \textbf{accuracy and reliability} of AI-generated responses across major Islamic knowledge domains, including the Qur'an, Hadith, Fiqh, ethics, pastoral guidance, and madhhab sensitivity, thereby addressing \textbf{RQ1}. Section~\ref{sec:rq2_authenticity} examines the \textbf{jurisprudential reasoning capabilities} of different AI models by comparing their citation practices, fiqh consistency, uncertainty recognition, and handling of scholarly disagreement, providing evidence for \textbf{RQ3}.

The analysis then investigates the \textbf{hallucination behaviour and citation reliability} of AI-generated Islamic responses in Section~\ref{sec:rq2_hallucination}, focusing on fabricated or incomplete citations, unverifiable references, and evidence transparency to address \textbf{RQ2}. Section~\ref{sec:rq2_sources} extends this analysis by examining the \textbf{knowledge provenance and source ecosystem} underlying AI-generated responses, identifying the dominant Islamic resources that influence generated content and discussing their implications for authenticity, diversity of scholarship, and trustworthiness. Finally, Section~\ref{sec:rq4_reliability} presents real-world case studies illustrating \textbf{reliability challenges in high-trust Islamic environments}, including factual errors and inconsistent retrieval behaviour, to evaluate the suitability of current generative AI systems for religious guidance, education, and Islamic research, thereby addressing \textbf{RQ4}. Collectively, these findings provide a comprehensive assessment of the strengths, limitations, and practical implications of using generative AI systems in Islamic knowledge domains.

\subsection{Accuracy and Reliability Across Islamic Knowledge Domains}
\label{sec:rq1_domains}

This subsection addresses \textbf{RQ1} by evaluating how accurately and authentically leading generative AI systems answer Islamic questions across the major domains of Qur'anic interpretation, Hadith explanation, Fiqh reasoning, ethical guidance, pastoral guidance, and Madhhab sensitivity. It also contributes to answering \textbf{RQ4} by assessing whether the observed level of performance is sufficient for deploying AI systems in high-trust Islamic environments such as education, religious guidance, and Islamic research.

\begin{figure}[htbp]
\centering
\includegraphics[width=0.75\textwidth]{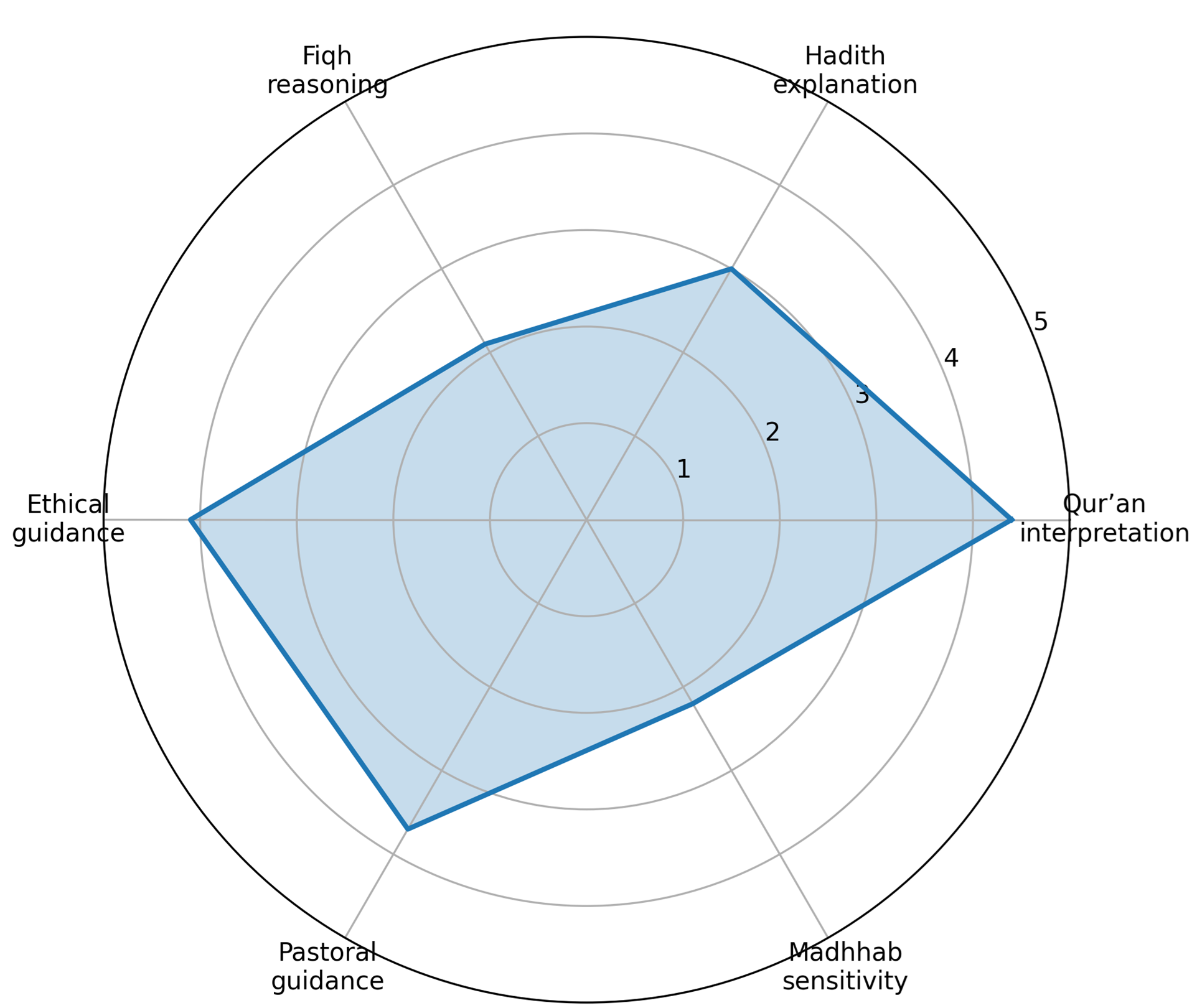}
\caption{Performance of the evaluated AI systems across six Islamic knowledge domains.}
\label{fig:islamic_domains}
\end{figure}

Figure~\ref{fig:islamic_domains} presents the comparative performance of the evaluated AI systems across six Islamic knowledge domains: Qur'anic interpretation, Hadith explanation, Fiqh reasoning, ethical guidance, pastoral guidance, and Madhhab sensitivity. The results demonstrate that AI capability is highly dependent on the complexity of the underlying religious knowledge. Domains supported by strong scholarly consensus generally achieve higher performance than those requiring advanced jurisprudential reasoning and interpretation.

The highest performance is observed in Qur'anic interpretation, achieving approximately 4.4 out of 5. This suggests that contemporary AI systems can generally explain Qur'anic verses, identify themes, and provide contextual interpretations with relatively high accuracy when compared with recognised Islamic references. Ethical guidance also demonstrates strong performance, scoring slightly above 4.0, reflecting the ability of AI systems to generate advice grounded in widely accepted Islamic moral principles. Similarly, pastoral guidance achieves a relatively high score of approximately 3.8, indicating that AI systems can often provide supportive responses to personal and spiritual questions where broad ethical principles are sufficient.

Performance decreases noticeably for Hadith explanation, which received a moderate score of approximately 3.0. Although AI systems frequently identify well-known Hadiths, limitations remain in accurately citing sources, recognising authentication status, explaining chains of narration, and providing sufficient contextual interpretation. The weakest performance is observed in Fiqh reasoning and Madhhab sensitivity, which achieved scores of approximately 2.1 and 2.3 respectively. These domains require sophisticated understanding of Islamic legal methodology, contextual reasoning, and legitimate scholarly disagreement. The lower scores indicate that current AI systems frequently struggle to distinguish between jurisprudential opinions, correctly represent different madhhabs, and explain the reasoning behind legal rulings.

\begin{figure}[htbp]
\centering
\includegraphics[width=0.80\textwidth]{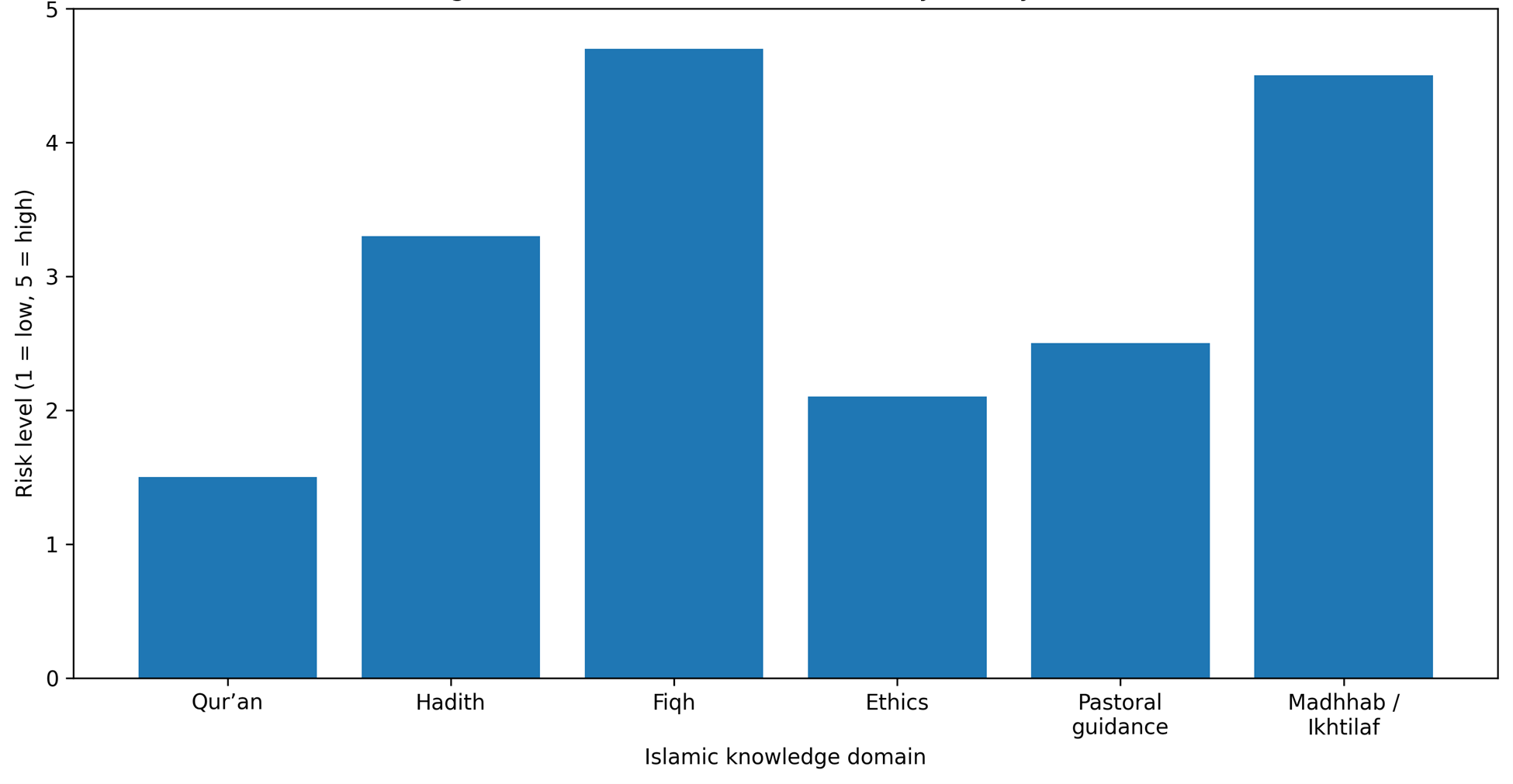}
\caption{Risk assessment of AI-generated responses across Islamic knowledge domains.}
\label{fig:islamic_risk_domains}
\end{figure}

Figure~\ref{fig:islamic_risk_domains} complements the performance analysis by evaluating the relative risk associated with relying on AI-generated responses across the same Islamic knowledge domains. The risk scale ranges from 1 (low risk) to 5 (high risk), where higher values represent a greater likelihood of inaccurate, incomplete, or potentially misleading religious guidance.

Consistent with the performance analysis, Qur'anic interpretation exhibits the lowest risk (approximately 1.5), followed by ethical guidance (approximately 2.1). These findings suggest that AI systems are comparatively reliable when explaining topics supported by extensive textual evidence and broad scholarly consensus. Moderate risk is observed for pastoral guidance (approximately 2.5) and Hadith explanation (approximately 3.3), primarily because these domains require contextual interpretation, accurate source attribution, and authentication of narrations.

The highest risk occurs within Fiqh and Madhhab/Ikhtilaf, with scores approaching 4.7 and 4.5 respectively. These domains involve complex jurisprudential reasoning, multiple valid scholarly opinions, and context-dependent legal rulings. AI systems frequently simplify legal discussions, overlook important qualifications, or fail to clearly distinguish between competing scholarly positions, thereby increasing the likelihood of incomplete or misleading guidance.

Taken together, Figures~\ref{fig:islamic_domains} and~\ref{fig:islamic_risk_domains} provide a comprehensive answer to \textbf{RQ1}. The results demonstrate that the accuracy and authenticity of generative AI systems vary considerably across Islamic knowledge domains, with the strongest performance observed in Qur'anic interpretation and ethical guidance, and the weakest performance in Fiqh reasoning and Madhhab-sensitive questions. These findings also contribute to \textbf{RQ4}, indicating that while current AI systems may serve as valuable tools for introductory Islamic education and general religious learning, they are not yet sufficiently reliable to be used independently for high-trust tasks involving jurisprudential reasoning, legal rulings, or scholarly interpretation without appropriate human oversight.

% updated 5.2

\subsection{Jurisprudential Reasoning and Model Consistency}
\label{sec:rq2_authenticity}

This subsection addresses \textbf{RQ3} by comparing how different generative AI systems handle Islamic jurisprudential reasoning, including fiqh consistency, citation practices, recognition of scholarly disagreement, and uncertainty handling. These characteristics are particularly important when evaluating whether AI systems can responsibly present multiple madhhab opinions, acknowledge ambiguity, and avoid presenting disputed rulings as universally accepted facts. Rather than focusing solely on linguistic quality, the analysis evaluates whether current generative AI systems provide sufficiently reliable citations, maintain consistent jurisprudential reasoning, and appropriately acknowledge scholarly uncertainty. These characteristics are fundamental for assessing whether AI-generated Islamic content can be independently verified and trusted in religious contexts.

\begin{figure}[htbp]
\centering
\includegraphics[width=0.85\textwidth]{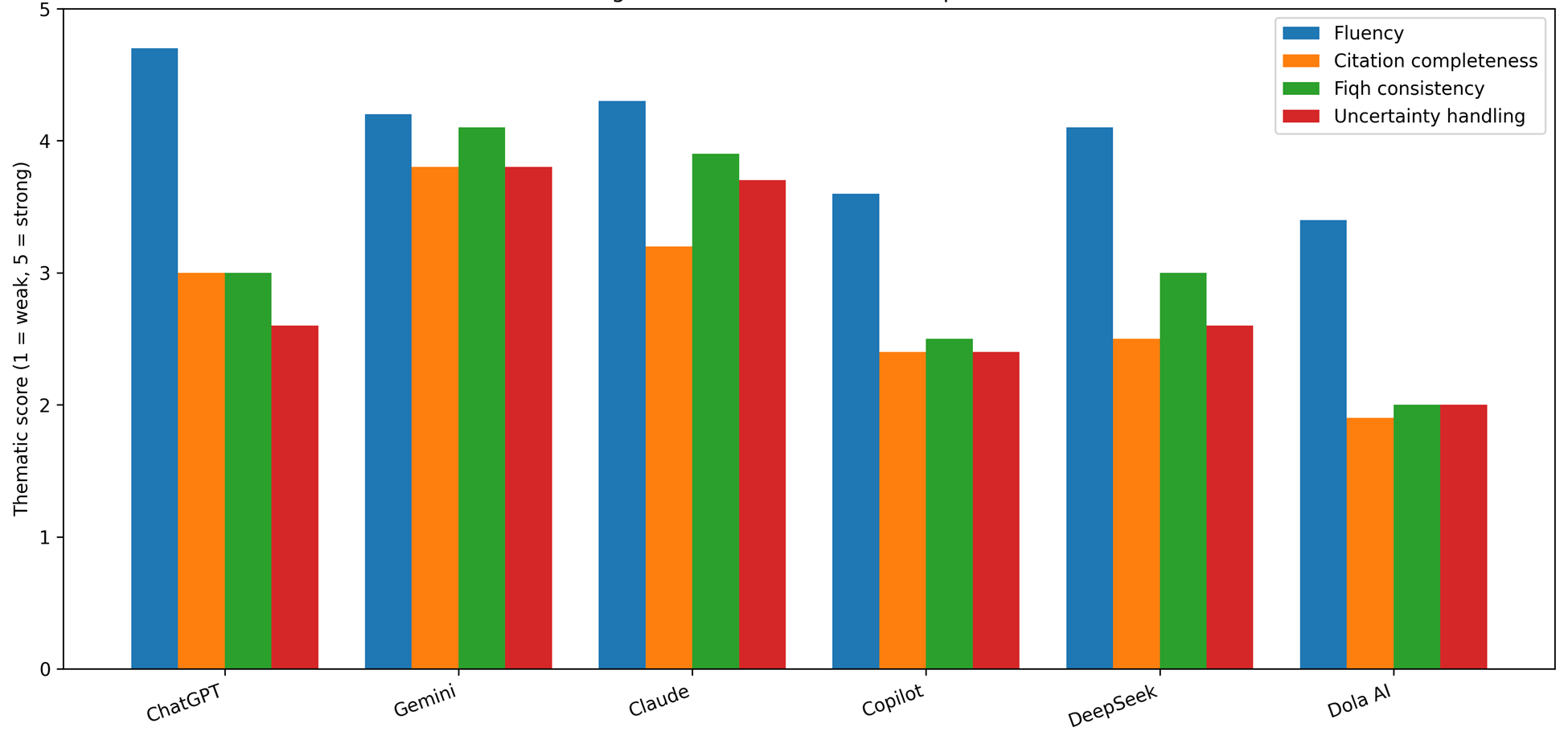}
\caption{Comparative thematic evaluation of AI systems across fluency, citation completeness, fiqh consistency, and uncertainty handling.}
\label{fig:thematic_evaluation}
\end{figure}

Figure~\ref{fig:thematic_evaluation} presents a comparative thematic evaluation of six AI systems  ChatGPT, Gemini, Claude, Copilot, DeepSeek, and Dola AI across four dimensions that directly influence the reasoning behaviour and reliability of Islamic responses: fluency, citation completeness, fiqh consistency, and uncertainty handling. The scores range from 1 (weak) to 5 (strong), where higher values indicate better performance within each evaluation criterion. While all systems produced fluent and coherent responses, considerably larger differences emerged in the dimensions associated with evidential quality and religious reliability.

Across all evaluated systems, fluency achieved the highest scores. ChatGPT obtained the highest fluency score (4.7), followed closely by Claude (4.3), Gemini (4.2), and DeepSeek (4.1). These results indicate that modern large language models are highly capable of generating natural, coherent, and persuasive Islamic explanations. However, fluency alone should not be interpreted as evidence of factual correctness or scholarly authenticity. The ability to produce well-written responses does not necessarily imply that the underlying religious evidence is complete, accurate, or appropriately referenced.

Greater variation is observed in citation completeness, which represents one of the most important indicators of evidence quality. Gemini achieved the highest score (3.8), followed by Claude (3.2) and ChatGPT (3.0), while DeepSeek, Copilot, and Dola AI demonstrated comparatively weaker citation practices. Across most systems, responses frequently omitted exact Hadith numbers, detailed source information, scholar names, or complete references, making independent verification difficult. Since Islamic scholarship places considerable importance on traceable evidence, these findings indicate that citation completeness remains a significant limitation of current generative AI systems.

Gemini and Claude demonstrated the greatest consistency when reasoning about Islamic jurisprudential issues. Both systems more frequently maintained internally coherent explanations across multiple fiqh scenarios and were more likely to recognise legitimate scholarly diversity. In contrast, ChatGPT, DeepSeek, Copilot and Dola AI exhibited greater variability, occasionally presenting simplified rulings or inconsistent reasoning when discussing issues involving multiple schools of thought. Although several systems were capable of providing reasonable fiqh explanations, maintaining coherent legal reasoning across questions involving different contexts, rulings, and scholarly opinions remains challenging.

A similar trend is observed for uncertainty handling, which evaluates the extent to which AI systems recognise ambiguity, acknowledge legitimate scholarly disagreement, and avoid presenting uncertain rulings as definitive conclusions. Gemini (3.8) and Claude (3.7) achieved the highest scores, indicating a greater tendency to recognise situations where multiple scholarly opinions exist. Other systems demonstrated weaker performance, often presenting simplified or overly confident responses despite the presence of recognised juristic disagreement. Since uncertainty and scholarly diversity are fundamental characteristics of many areas of Islamic jurisprudence, responsible acknowledgement of these limitations is essential for trustworthy AI-generated religious guidance.

Overall, Figure~\ref{fig:thematic_evaluation} provides a direct answer to \textbf{RQ3}. The results demonstrate that although all evaluated AI systems produce fluent Islamic responses, substantial differences exist in their ability to reason consistently about jurisprudential issues, recognise legitimate scholarly disagreement, and appropriately communicate uncertainty. Gemini and Claude consistently demonstrated stronger performance across fiqh consistency and uncertainty handling, whereas other systems exhibited greater variability when reasoning about complex Islamic legal questions. These findings suggest that responsible handling of jurisprudential diversity remains an important challenge for current generative AI systems.

% 5.2 ends

% 5.3 starts 

\subsection{Hallucinations and Citation Reliability in AI-Generated Islamic Responses}
\label{sec:rq2_hallucination}

This subsection further addresses \textbf{RQ2} by examining the types of hallucinations and citation-related deficiencies observed in AI-generated Islamic responses. Beyond evaluating overall evidence quality, the analysis investigates whether AI systems fabricate or omit references, provide unverifiable religious claims, merge distinct scholarly opinions, or present incomplete citations that reduce the transparency and authenticity of religious guidance.

\begin{table}[H]
\centering
\caption{Observed Hallucination-Related Issues in AI-Generated Islamic Responses}
\label{tab:hallucination_issues}
\renewcommand{\arraystretch}{1.2}

\begin{tabular}{|p{3.8cm}|p{2.5cm}|p{10cm}|}
\hline
\textbf{Hallucination Issue} &
\textbf{Observation} &
\textbf{Example} \\
\hline

Missing Hadith Citation &
Common &
Hadiths were mentioned without providing the exact book, chapter, or Hadith number, limiting independent verification. \\
\hline

General References Without Attribution &
Common &
Responses used phrases such as ``The Prophet said'', ``A Hadith states'', or ``Scholars have said'' without identifying the source. Frequently observed in ChatGPT and Copilot responses. \\
\hline

Unverified Religious Claims &
Present &
Religious rulings and recommendations were occasionally provided without supporting Qur'anic verses, authentic Hadiths, or scholarly references. \\
\hline

Incomplete Source Attribution &
Frequent &
Statements such as ``Reported in Sahih al-Bukhari'' or ``Reported in Sahih Muslim'' were provided without exact narration references. \\
\hline

Unnamed Scholarly Opinions &
Common &
Responses referred to ``scholars'' or ``Islamic scholars'' without identifying the scholar, madhhab, or source of the opinion. \\
\hline

Merged Scholarly Views &
Common &
Different scholarly opinions were combined into a single response without distinguishing between the individual viewpoints or legal schools. \\
\hline

Overconfident Responses &
Occasional &
Definitive conclusions were presented despite recognised scholarly disagreement, particularly in contemporary fiqh topics such as insurance. \\
\hline

Potential Citation Hallucination &
Rare &
References appeared authentic but lacked sufficient information for independent verification or source tracing. \\
\hline

\end{tabular}
\end{table}

Table~\ref{tab:hallucination_issues} summarises the principal hallucination-related issues identified throughout the evaluation. Rather than generating entirely fabricated Islamic content, the most frequently observed problems involved citation incompleteness, insufficient scholarly attribution, and inadequate evidence supporting religious claims. Missing Hadith identifiers, unnamed scholars, incomplete references, and overconfident presentation of disputed jurisprudential opinions occurred considerably more often than outright factual fabrication. These findings indicate that the dominant reliability challenge in current generative AI systems lies in evidential transparency rather than purely factual accuracy.

\begin{figure*}[htbp]
\centering
\includegraphics[width=\textwidth]{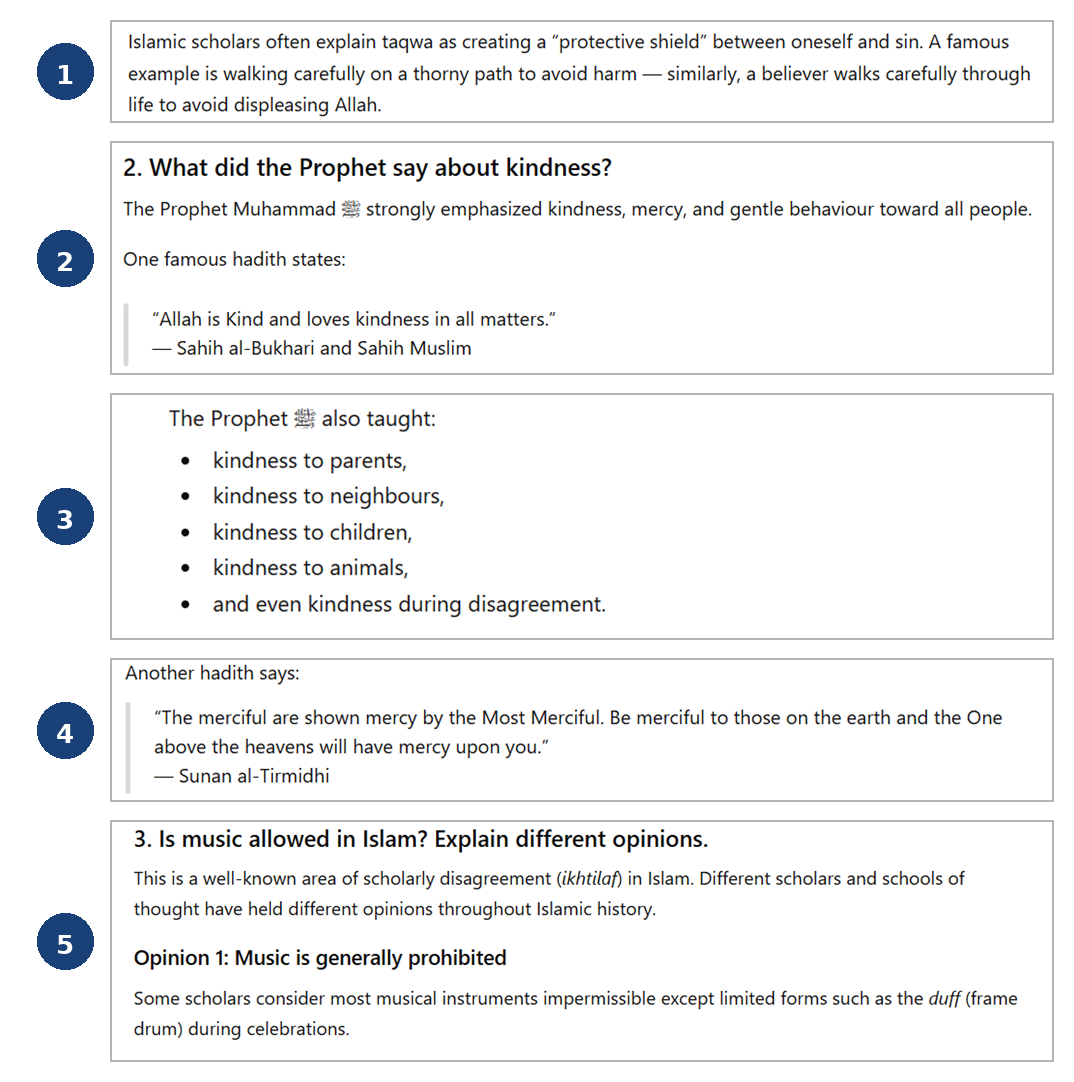}
\caption{Examples of AI-generated Islamic responses exhibiting generic explanations and incomplete source attribution.}
\label{fig:generic_references_examples}
\end{figure*}

Figure~\ref{fig:generic_references_examples} provides representative examples illustrating the hallucination patterns summarised in Table~\ref{tab:hallucination_issues}. Across multiple Islamic topics, including Taqwa, kindness, mercy, and music in Islam, the AI systems generally produced linguistically fluent and contextually reasonable explanations. However, the responses frequently omitted essential scholarly evidence required for independent verification. Common observations include the absence of exact Hadith numbers, missing book and chapter references, unidentified scholars, and generic statements such as ``The Prophet said'' or ``scholars have stated'' without adequate attribution. Although these responses often appear authoritative, the absence of precise references substantially reduces their transparency, traceability, and scholarly reliability.

The figure further demonstrates that AI systems frequently oversimplify complex jurisprudential discussions. For topics involving recognised scholarly disagreement, such as the permissibility of music, multiple opinions are often summarised without identifying the respective scholars, schools of thought, or evidential basis supporting each position. Likewise, Hadith-related responses commonly reference collections such as \textit{Sahih al-Bukhari}, \textit{Sahih Muslim}, or \textit{Sunan al-Tirmidhi} without providing exact narration numbers, preventing users from independently verifying authenticity or context. While these observations do not necessarily constitute factual hallucinations, they represent a significant form of citation hallucination and evidence incompleteness that can undermine user confidence and scholarly accountability.

Overall, this analysis provides a direct answer to \textbf{RQ2}. The findings indicate that hallucinations within AI-generated Islamic responses are more commonly associated with incomplete citations, missing scholarly attribution, unverifiable references, and overconfident presentation of disputed religious issues than with completely fabricated Islamic content. Consequently, although generative AI systems can provide useful introductory explanations, their responses should be independently verified using primary Islamic sources and recognised scholarly authorities before being relied upon for religious guidance, academic research, or jurisprudential decision making.

% 5.3 ends

% 5.4 starts

\subsection{Knowledge Provenance and Source Ecosystem}
\label{sec:rq2_sources}

This subsection addresses \textbf{RQ2} by investigating the provenance of AI-generated Islamic knowledge and identifying the dominant online sources that influence generated responses. It also contributes to \textbf{RQ4} by examining how reliance on a relatively small ecosystem of digital Islamic resources affects the suitability of generative AI systems for high-trust environments such as Islamic education, religious guidance, and scholarly research.

The analysis indicates that generative AI systems frequently rely on a relatively small and highly centralised ecosystem of Islamic websites, databases, and educational platforms when generating Islamic responses. Rather than drawing uniformly from the broad spectrum of Islamic scholarship, AI-generated answers consistently reference a limited number of highly indexed online resources. This finding suggests that the quality, diversity, and authenticity of AI-generated Islamic knowledge are strongly influenced by the visibility and accessibility of digital Islamic content.

For Qur'anic content, platforms such as \textit{Quran.com}, \textit{Tanzil}, and \textit{Altafsir} appear to be the dominant retrieval sources because of their extensive indexing, multilingual availability, and widespread adoption across Islamic educational websites. Consequently, AI-generated Qur'anic explanations often exhibit highly consistent wording, standardised translations, and similar thematic interpretations across different AI systems.

For Hadith-related questions, AI systems commonly reference digital repositories such as \textit{Sunnah.com} together with major Hadith collections including \textit{Sahih Muslim}, \textit{Sunan Abi Dawud}, and \textit{Jami' at-Tirmidhi}. However, despite frequently mentioning these collections, the analysis reveals a recurring limitation: AI-generated responses often omit essential scholarly information such as exact Hadith numbers, authenticity classifications, chains of narration, and complete citation details. This finding suggests that current AI systems prioritise conversational fluency over scholarly precision, reducing the transparency and verifiability of religious evidence.

The analysis further demonstrates the strong influence of classical tafsir literature and contemporary Islamic educational websites on AI-generated explanations. Topics including Taqwa, patience, charity, and other foundational Islamic concepts frequently resemble interpretations found in \textit{Tafsir Ibn Kathir}, \textit{Maarif-ul-Quran}, and the writings of scholars such as Ibn Kathir, Al-Qurtubi, and Abul Ala Maududi. Likewise, responses addressing contemporary Islamic issues frequently align with content published by \textit{IslamQA}, \textit{SeekersGuidance}, and the \textit{Yaqeen Institute}, reflecting the strong online presence of these organisations.

The Islamic knowledge sources most frequently identified during the evaluation are summarised in Table~\ref{tab:islamic_sources}.

\begin{table}[H]
\centering
\caption{Islamic Knowledge Sources Commonly Referenced in AI-Generated Responses}
\label{tab:islamic_sources}
\begin{tabular}{|p{5cm}|p{8cm}|}
\hline
\textbf{Institution / Platform} & \textbf{Official URL} \\
\hline
Quran.com & https://quran.com \\
\hline
Tanzil Project & https://tanzil.net \\
\hline
Altafsir & https://altafsir.com \\
\hline
Sunnah.com & https://sunnah.com \\
\hline
IslamQA & https://islamqa.info \\
\hline
SeekersGuidance & https://seekersguidance.org \\
\hline
Yaqeen Institute & https://yaqeeninstitute.org \\
\hline
Abu Amina Elias & https://www.abuaminaelias.com \\
\hline
Dar al-Ifta Egypt & https://www.dar-alifta.org \\
\hline
Al-Azhar University & https://www.azhar.eg \\
\hline
IslamWeb & https://www.islamweb.net \\
\hline
Muslim Matters & https://muslimmatters.org \\
\hline
Virtual Mosque & https://www.virtualmosque.com \\
\hline
Bayyinah Institute & https://bayyinahtv.com \\
\hline
Cambridge Muslim College & https://cambridgemuslimcollege.ac.uk \\
\hline
Al-Madina Institute & https://almadinainstitute.org \\
\hline
Reddit Islam Discussions & https://www.reddit.com/r/islam/ \\
\hline
Wikipedia Islamic Topics & https://www.wikipedia.org \\
\hline
\end{tabular}
\end{table}

A particularly important finding is the phenomenon of \textit{source ecosystem centralisation}, whereby a relatively small number of highly visible online resources disproportionately influence AI-generated Islamic responses. This concentration leads to repeated use of similar interpretations, standardised wording, and reduced exposure to the broader diversity of Islamic scholarship. Consequently, minority scholarly opinions, regional traditions, and less-digitised Islamic literature are underrepresented. In areas involving juristic disagreement, such as music, financial transactions, or contemporary ethical issues, AI systems frequently simplify complex scholarly discussions or fail to clearly distinguish between competing schools of thought.

Overall, this analysis provides additional evidence for both \textbf{RQ2} and \textbf{RQ4}. The findings demonstrate that the authenticity of AI-generated Islamic responses is shaped not only by the capabilities of the language model itself, but also by the limited ecosystem of online sources from which knowledge is retrieved or learned. While this source ecosystem enables AI systems to provide useful introductory explanations and rapidly surface relevant Islamic information, reliance on a relatively narrow collection of digital resources may also contribute to incomplete citations, simplified interpretations, and insufficient representation of scholarly diversity. Consequently, AI-generated responses should be treated as informational assistance rather than authoritative religious guidance, reinforcing the continuing importance of verification using primary Islamic sources and qualified scholars in high-trust educational and religious environments.

% 5.4 ends

% 5.5 starts

\subsection{Reliability Challenges in High-Trust Islamic Environments}
\label{sec:rq4_reliability}

This subsection primarily addresses \textbf{RQ4} by examining whether current generative AI systems are sufficiently reliable for high-trust Islamic environments such as religious guidance, Islamic education, and scholarly research. It also provides additional evidence for \textbf{RQ2} by presenting concrete examples of factual inaccuracies and retrieval inconsistencies that may not be apparent through aggregate quantitative evaluation alone.

\begin{figure}[t]
\centering
\includegraphics[width=\columnwidth]{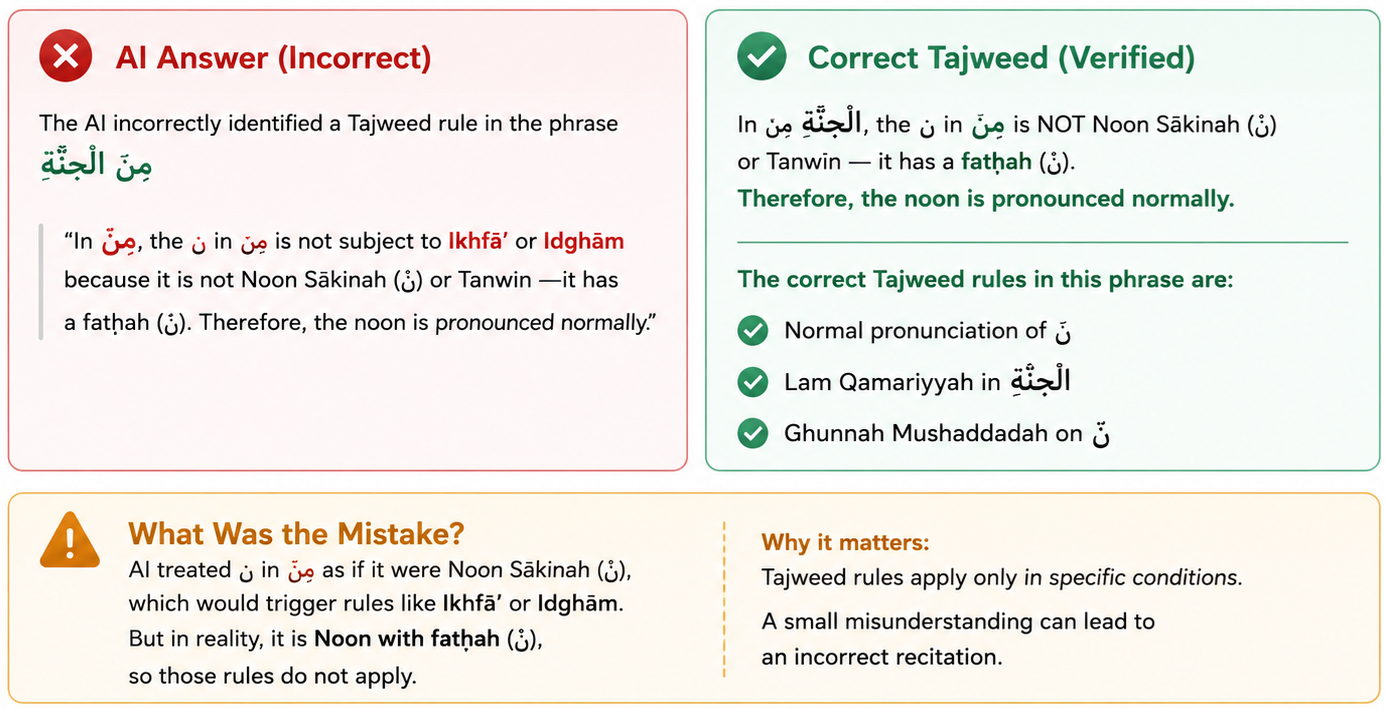}
\caption{Example illustrating a hallucination in a generative AI explanation of Tajweed rules.}
\label{fig:tajweed_ai}
\end{figure}

Figure~\ref{fig:tajweed_ai} demonstrates that generative AI systems can occasionally produce technically incorrect religious explanations despite presenting them with a high degree of confidence. In this example, the AI incorrectly classified the Arabic letter \textit{Noon} in a Tajweed rule, leading to an incorrect explanation of Qur'anic recitation. Although the mistake appears relatively minor, Tajweed is governed by precise linguistic and recitation rules where even small errors may alter the correct explanation of Qur'anic pronunciation. This example illustrates an important characteristic of generative AI systems: fluent and confident responses do not necessarily guarantee technical correctness. In specialised religious domains, users without sufficient background knowledge may find it difficult to recognise such errors, increasing the risk of misinformation being accepted as authentic Islamic guidance.

\begin{figure}[t]
\centering
\includegraphics[width=\linewidth]{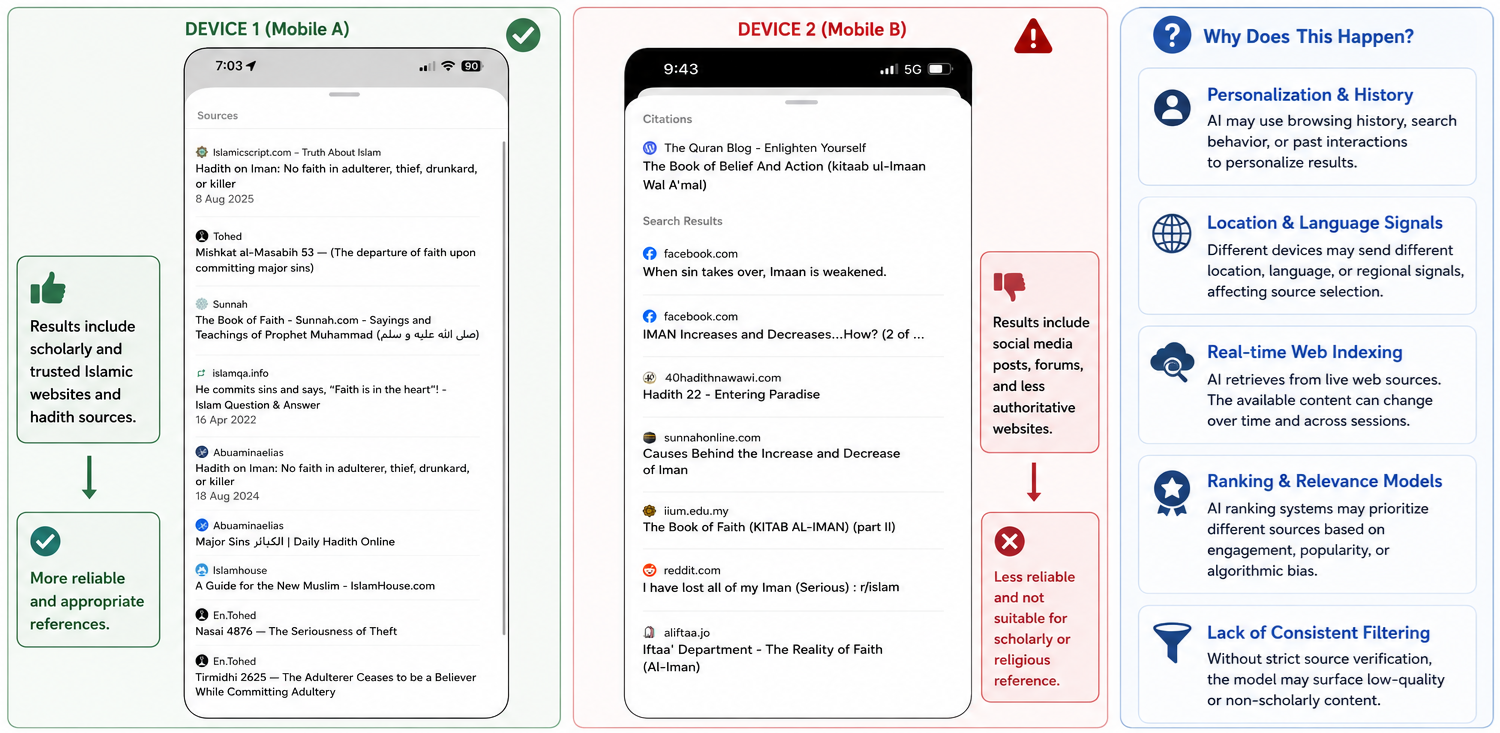}
\caption{Generative AI provides different sources for the same query across different devices, demonstrating variability in AI-generated reference retrieval.}
\label{fig:ai_different_sources}
\end{figure}

Figure~\ref{fig:ai_different_sources} demonstrates a second reliability challenge: identical queries submitted through different devices may produce responses supported by substantially different reference sources. In the presented example, the same question concerning the concept of \textit{Imaan} in Islam was submitted using two mobile devices. One device primarily returned references from recognised Islamic educational websites and authenticated Hadith collections, whereas the other produced responses supported by a mixture of social media platforms, online discussion forums, and less authoritative websites. Although the generated answers were broadly similar, the underlying evidence varied considerably in both quality and authenticity.

Such variations may arise from several factors including personalised user history, language preferences, regional search settings, retrieval-augmented search mechanisms, real-time web indexing, and continuously evolving ranking algorithms. Consequently, generative AI should not be viewed as producing deterministic outputs for knowledge retrieval. Instead, the retrieval process may differ depending on the user context, resulting in varying levels of source reliability even for identical questions.

Taken together, Figures~\ref{fig:tajweed_ai} and~\ref{fig:ai_different_sources} provide practical evidence supporting the findings presented throughout this study. While the quantitative analyses demonstrated limitations in citation completeness, jurisprudential reasoning, and evidence quality, these case studies illustrate how such limitations manifest during real-world use. They show that AI systems may generate technically incorrect religious explanations or rely on inconsistent and potentially less authoritative reference sources despite producing fluent and convincing responses. These observations directly answer \textbf{RQ4}, indicating that although current generative AI systems can effectively support introductory Islamic learning and information discovery, they should not be regarded as authoritative sources for religious rulings, scholarly interpretation, or academic research without careful verification by qualified scholars and authenticated primary sources.

\section{Recommendations and Future Work}
\label{sec:recommendations_future_work}

Based on the findings of this study, generative AI should be used in Islamic research as an assistive tool rather than as an authoritative source of religious knowledge. AI systems may be useful for generating introductory explanations, identifying possible themes, comparing broad viewpoints, drafting summaries, and supporting early-stage literature exploration. However, AI-generated Islamic responses should not be accepted without verification, particularly when they involve Qur'anic interpretation, Hadith authentication, Fiqh rulings, Madhhab-specific positions, or contemporary religious issues involving scholarly disagreement.

For Islamic research, users should apply a verification-first approach. Any Qur'anic reference generated by AI should be checked against the Arabic text, recognised translations, and established tafsir sources. Hadith references should be verified using authentic Hadith collections, exact narration numbers, grading, and contextual explanation. Fiqh-related responses should be checked against recognised scholars, reliable fatwa institutions, and the relevant Madhhab position. Where multiple scholarly opinions exist, researchers should avoid relying on a single AI-generated answer and should instead identify the evidential basis, legal reasoning, and school-of-thought context behind each view.

The findings also suggest that users should be cautious of fluent but weakly sourced responses. A response may appear clear, balanced, and persuasive while still containing incomplete citations, missing Hadith identifiers, unnamed scholarly opinions, or oversimplified jurisprudential reasoning. Therefore, AI-generated Islamic content should be treated as a starting point for investigation, not as final evidence. In educational settings, students may use AI to support understanding, but they should be required to cite primary Islamic sources and explain how they verified AI-generated claims. In scholarly research, AI outputs should not be cited as religious authority unless the underlying sources have been independently traced and validated.

This study also recommends that AI developers working on Islamic knowledge systems should improve source transparency, citation fidelity, and responsible abstention. AI systems should provide exact Qur'anic references, Hadith numbers, authentication status, scholar names, Madhhab context, and clear uncertainty statements where appropriate. For disputed issues, systems should avoid presenting one view as universal unless scholarly consensus exists. Instead, they should explain that legitimate disagreement exists and identify the relevant schools of thought or recognised scholarly positions.

Future work should extend this study using a much larger dataset of Islamic questions across a broader range of domains, languages, regions, Madhhabs, and difficulty levels. While the present study used fifty realistic open-ended Islamic questions, future research should experiment with hundreds or thousands of prompts covering Qur'an, Hadith, Fiqh, Aqeedah, Seerah, Islamic finance, family law, inheritance, contemporary bioethics, and pastoral counselling. A larger dataset would enable more robust statistical analysis of hallucination rates, citation accuracy, Madhhab sensitivity, abstention behaviour, and model consistency.

Future research should also examine multilingual Islamic AI performance, particularly in Arabic, Urdu, Malay, Turkish, Persian, Indonesian, and other languages commonly used in Islamic scholarship and Muslim communities. Since many authoritative Islamic sources are not equally represented in English-language digital datasets, multilingual evaluation may reveal different patterns of accuracy, source dependence, and jurisprudential interpretation. Further work should also compare retrieval-augmented Islamic AI systems with general-purpose AI systems to determine whether curated Islamic knowledge bases can reduce hallucinations, improve citation precision, and support more responsible religious guidance.

Finally, future studies should investigate geographic and retrieval-dependent variation in AI-generated Islamic responses. The present study observed that responses collected from Australia and the United Kingdom were often semantically similar but differed in wording, citation completeness, retrieved sources, and supporting references. Larger-scale experiments should systematically test whether identical prompts submitted from different countries, devices, languages, or user profiles produce different Islamic references or interpretations. Such work is important for improving reproducibility, transparency, and trust in AI-assisted Islamic education and research.

\section{Conclusion}

This paper presented a comprehensive evaluation of the reliability, authenticity, and interpretive quality of leading generative AI systems within Islamic knowledge domains. Using a realistic dataset of fifty open-ended Islamic questions and a mixed-method evaluation framework, the study examined AI-generated responses across Qur'anic interpretation, Hadith explanation, Fiqh reasoning, ethical guidance, pastoral advice, and Madhhab-sensitive topics. To investigate the consistency of AI-generated religious information across different retrieval environments, responses were collected and compared from both the United Kingdom and Australia.

The findings directly address the four research questions. For \textbf{RQ1}, the evaluation demonstrated that current AI systems achieve comparatively high accuracy in domains characterised by broad scholarly consensus, particularly Qur'anic interpretation and ethical guidance, while performance decreases substantially for jurisprudential reasoning and Madhhab-sensitive questions. For \textbf{RQ2}, the study identified recurring hallucination-related issues, including incomplete citations, missing Hadith identifiers, unverifiable references, unnamed scholarly opinions, and occasional fabricated or insufficiently supported religious claims. These findings indicate that fluent AI-generated responses cannot be assumed to possess equivalent scholarly authenticity.

Regarding \textbf{RQ3}, the comparative analysis revealed considerable variation among AI systems in their handling of jurisprudential reasoning, uncertainty recognition, and scholarly disagreement. While some models demonstrated stronger consistency and greater willingness to acknowledge multiple valid opinions, others tended to present simplified or overconfident rulings in areas where legitimate differences of opinion exist. Finally, addressing \textbf{RQ4}, the study showed that current generative AI systems remain unsuitable as standalone authorities for high-trust Islamic environments. Their responses are strongly influenced by a relatively small ecosystem of online Islamic resources and may vary in reliability, citation quality, interpretive completeness, and supporting evidence.

An additional contribution of this study is the investigation of AI behaviour across different geographic retrieval environments. By comparing responses generated from the United Kingdom and Australia, the study found that although AI systems generally produced semantically similar answers, they did not consistently provide identical supporting evidence. Variations were observed in wording, citation completeness, source attribution, and the references presented for identical Islamic queries. These findings suggest that generative AI systems may exhibit retrieval-dependent behaviour, whereby the same prompt can produce different supporting evidence depending on the retrieval environment or geographic access conditions. This observation has important implications for reproducibility, transparency, and trust in AI-assisted religious guidance.

Overall, the findings suggest that generative AI should be viewed as an assistive educational technology rather than an authoritative religious advisor. While these systems can effectively support introductory Islamic learning and information discovery, their outputs remain susceptible to hallucinations, incomplete citations, inconsistent jurisprudential reasoning, retrieval-dependent behavioural variations, and source attribution issues. Consequently, AI-generated Islamic information should always be verified against authenticated primary sources, recognised scholarly literature, and qualified Islamic scholars before being relied upon for religious guidance, education, or academic research. Future research should investigate retrieval-augmented Islamic AI, benchmark datasets covering broader schools of thought, multilingual evaluation, geographic consistency of AI-generated responses, automated source verification, and methods for improving citation fidelity, responsible abstention, and transparency in AI-assisted religious guidance.

\section*{Acknowledgements}

This research is done under Islamic Research Center, Sydney.I would like to acknowledge the valuable contributions of  \textbf{[Dr. Muhammad Sajjad Akbar]} (\textit{University of Sydney, Australia} [contact email of the organizer: sajjad.akr1@gmail.com]) 
\textbf{[Dr Zawar Hussain]} (\textit{Macquarie University})
\textbf{[Imran Afzal Khan]} (\textit{University of New South Wales})
\textbf{[Dr Muhammad Qasim Malik]} (\textit{General Dentist, Sydney})
\textbf{[Dr Muhammad Ikram]} (\textit{Macquarie University})
\textbf{[Ammar]} (\textit{Sydney})
\textbf{[Dr Ikram Asghar]} (\textit{Teesside University, UK})
\textbf{[Dr Saqib Shamim]} (\textit{Queen Mary University of London})
\textbf{[Dr Hammad Nazir]} (\textit{University of South Wales, UK})
\textbf{[Dr Rushan Arshad]} (\textit{Bournemouth University, UK})
\textbf{[Dr Dr Rehan Zia ]} (\textit{Bournemouth University, UK})
\textbf{[Dr Jawwad Latif]} (\textit{Bournemouth University, UK})
\textbf{[Faisal Malik]} (\textit{UK})
\textbf{[Dr Mohammad Polash]} (\textit{University of Sydney, Australia})
\textbf{[Dr Imdad Ullah]} (\textit{University of Sydney, Australia})
\textbf{[Abid Tariq Sheikh]} (\textit{Sydney}) 
for their assistance with the collection of AI-generated responses, data organisation, and independent review of the evaluation results. Their careful review and constructive feedback contributed to improving the quality, consistency, and reliability of the study.

\bibliographystyle{apalike}
\bibliography{reference}

@misc{microsoft2025globalai,
  author       = {{Microsoft AI Economy Institute}},
  title        = {Global AI Adoption in 2025: A Widening Digital Divide},
  year         = {2026},
  howpublished = {\url{https://www.microsoft.com/en-us/corporate-responsibility/topics/ai-economy-institute/reports/global-ai-adoption-2025/}},
  note         = {Accessed: 2026-05-09}
}

@misc{eurostat2025genai,
  author       = {{Eurostat}},
  title        = {32.7\% of EU People Used Generative AI Tools in 2025},
  year         = {2025},
  howpublished = {\url{https://ec.europa.eu/eurostat/web/products-eurostat-news/w/ddn-20251216-3}},
  note         = {Accessed: 2026-05-09}
}

@misc{hepi2025survey,
  author       = {Freeman, James and Hall, Rachel and Pownall, Madeleine},
  title        = {Student Generative AI Survey 2025},
  year         = {2025},
  institution  = {Higher Education Policy Institute (HEPI)},
  howpublished = {\url{https://www.hepi.ac.uk/reports/student-generative-ai-survey-2025/}},
  note         = {Accessed: 2026-05-09}
}

@misc{microsoft2025education,
  author       = {{Microsoft}},
  title        = {2025 Microsoft AI in Education Report},
  year         = {2025},
  howpublished = {\url{https://cdn-dynmedia-1.microsoft.com/is/content/microsoftcorp/microsoft/bade/documents/products-and-services/en-us/education/2025-Microsoft-AI-in-Education-Report.pdf}},
  note         = {Accessed: 2026-05-09}
}

@article{jacovi2025facts,
  title={The FACTS Grounding Leaderboard: Benchmarking LLMs' Ability to Ground Responses to Long-Form Input},
  author={Jacovi, Alon and Wang, Andrew and Alberti, Chris and Tao, Connie and Lipovetz, Jon and Olszewska, Kate and Haas, Lukas and Liu, Michelle and Keating, Nate and Bloniarz, Adam and others},
  journal={arXiv preprint arXiv:2501.03200},
  year={2025}
}

@article{wei2024measuring,
  title={Measuring short-form factuality in large language models},
  author={Wei, Jason and Karina, Nguyen and Chung, Hyung Won and Jiao, Yunxin Joy and Papay, Spencer and Glaese, Amelia and Schulman, John and Fedus, William},
  journal={arXiv preprint arXiv:2411.04368},
  year={2024}
}

@article{linardon2025influence,
  title={Influence of topic familiarity and prompt specificity on citation fabrication in mental health research using large language models: experimental study},
  author={Linardon, Jake and Jarman, Hannah K and McClure, Zoe and Anderson, Cleo and Liu, Claudia and Messer, Mariel},
  journal={JMIR Mental Health},
  volume={12},
  pages={e80371},
  year={2025},
  publisher={JMIR Publications Toronto, Canada}
}

@article{magesh2025hallucination,
  title={Hallucination-free? Assessing the reliability of leading AI legal research tools},
  author={Magesh, Varun and Surani, Faiz and Dahl, Matthew and Suzgun, Mirac and Manning, Christopher D and Ho, Daniel E},
  journal={Journal of empirical legal studies},
  volume={22},
  number={2},
  pages={216--242},
  year={2025},
  publisher={Wiley Online Library}
}

@article{abdelaal2026islamicmmlu,
  title={IslamicMMLU: A Benchmark for Evaluating LLMs on Islamic Knowledge},
  author={Abdelaal, Ali and Haffar, Mohammed Nader Al and Fawzi, Mahmoud and Magdy, Walid},
  journal={arXiv preprint arXiv:2603.23750},
  year={2026}
}

@inproceedings{atif2025sacred,
  title={Sacred or Synthetic? Evaluating LLM Reliability and Abstention for Religious Questions},
  author={Atif, Farah and Askarbekuly, Nursultan and Darwish, Kareem and Choudhury, Monojit},
  booktitle={Proceedings of the AAAI/ACM Conference on AI, Ethics, and Society},
  volume={8},
  number={1},
  pages={217--226},
  year={2025}
}

@inproceedings{lahmar2025islamtrust,
  title={IslamTrust: A Benchmark for LLMs Alignment with Islamic Values},
  author={Lahmar, Abderraouf and Arafat, Md Easin and Farou, Zakarya and Mahmud, Mufti},
  booktitle={5th Muslims in ML Workshop co-located with NeurIPS 2025},
  year={2025}
}

@inproceedings{bouchekif2025assessing,
  title={Assessing large language models on islamic legal reasoning: Evidence from inheritance law evaluation},
  author={Bouchekif, Abdessalam and Rashwani, Samer and Sbahi, Heba and Gaben, Shahd and Al Khatib, Mutaz and Ghaly, Mohammed},
  booktitle={Proceedings of The Third Arabic Natural Language Processing Conference},
  pages={246--257},
  year={2025}
}

\end{document}